\documentclass[letterpaper, 10 pt, conference]{ieeeconf}

\IEEEoverridecommandlockouts

\usepackage{times}

\usepackage{cite}

\usepackage{multicol}
\usepackage[table]{xcolor}
\definecolor{cvprblue}{rgb}{0.21,0.49,0.74}
\usepackage[bookmarks=true,pagebackref,breaklinks,colorlinks,allcolors=cvprblue]{hyperref}

\usepackage{amsmath}
\usepackage{amssymb}
\usepackage{graphicx}
\usepackage{booktabs}
\usepackage{multirow}
\usepackage{cuted}
\usepackage{capt-of}
\usepackage{algorithm}
\usepackage{algpseudocode}
\usepackage{gensymb}
\usepackage{makecell}
\usepackage{pifont}
\newcommand{\cmark}{\ding{51}}
\newcommand{\xmark}{\ding{55}}
\usepackage{subcaption}
\usepackage{dsfont}
\usepackage{float}
\usepackage[accsupp]{axessibility}
\usepackage{tabularx}
\usepackage{array}
\makeatletter

\providecommand{\@mkboth}[2]{}
\makeatother
\usepackage{etoc}
\usepackage{etoolbox}
\newcommand{\method}{ULTRA}

\newcommand{\rebuttal}[1]{\textcolor{black}{#1}}

\newcolumntype{Y}{>{\raggedright\arraybackslash}X}

\begin{document}
\etocdepthtag.toc{mtchapter}

\title{\LARGE \bf
ULTRA: Unified Multimodal Control for Autonomous Humanoid Whole-Body Loco-Manipulation
}

\author{Xialin He$^{\dag}$ \quad Sirui Xu$^{\dag}$ \quad Xinyao Li \quad Runpei Dong \quad Liuyu Bian \\ 
Yu-Xiong Wang$^{\ddag}$ \quad
Liang-Yan Gui$^{\ddag}$\\
\small{University of Illinois Urbana-Champaign}\\
\small{$^{\dag}$Equal Contribution \quad $^{\ddag}$Equal Advising}\\
\small\url{https://ultrahumanoid.github.io/}}

\maketitle
\thispagestyle{empty}
\pagestyle{empty}

\begin{strip}\centering
\vspace{-3.6em}
\includegraphics[width=\textwidth]{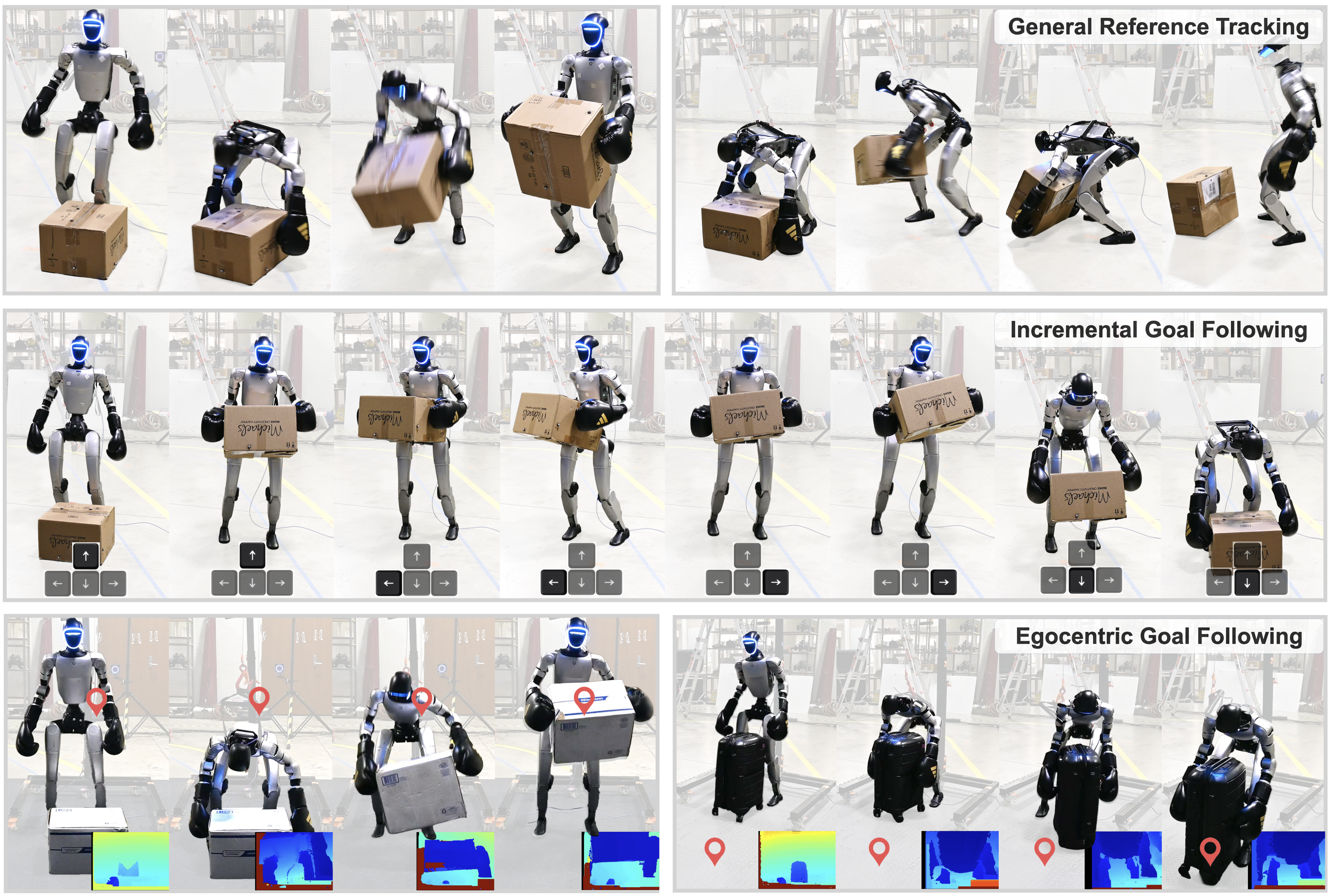}
\captionof{figure}{ULTRA is an \textbf{all-in-one} controller for humanoid loco-manipulation that supports: \textbf{Top:} dense motion tracking from \rebuttal{different} references; \textbf{Middle:} \textit{fine-grained control} from operator commands; \textbf{Bottom:} \textit{long-horizon goal} following with \textit{egocentric perception}. It demonstrates autonomous whole-body behavior without relying on test-time motion references under real sensing. We provide additional examples in our project \href{https://ultrahumanoid.github.io/}{webpage}.\label{fig:teaser}}
\vspace{-1em}
\end{strip}

\begin{abstract}
Achieving autonomous and versatile whole-body loco-manipulation remains a central barrier to making humanoids practically useful. Yet existing approaches are fundamentally constrained: retargeted data are often scarce or low-quality; methods struggle to scale to large skill repertoires; and, most importantly, they rely on tracking predefined motion references rather than generating behavior from perception and high-level task specifications.
To address these limitations, we propose ULTRA, a unified framework with two key components. First, we introduce a physics-driven neural retargeting algorithm that translates large-scale motion capture to humanoid embodiments while preserving physical plausibility for contact-rich interactions. Second, we learn a unified multimodal controller that supports both dense references and sparse task specifications, under sensing ranging from accurate motion-capture state to noisy egocentric visual inputs. We distill a universal tracking policy into this controller, compress motor skills into a compact latent space, and apply reinforcement learning finetuning to expand coverage and improve robustness under out-of-distribution scenarios. This enables coordinated whole-body behavior from sparse intent without test-time reference motions.
We evaluate ULTRA in simulation and on a real Unitree G1 humanoid. Results show that ULTRA generalizes to autonomous, goal-conditioned whole-body loco-manipulation from egocentric perception, consistently outperforming tracking-only baselines with limited skills.

\end{abstract}

\section{Introduction}

Real-world loco-manipulation requires autonomy beyond replaying fixed reference motions: a humanoid must span a continuum from dense motion references to sparse task goals, and from accurate state estimation to onboard sensing. Yet many controllers treat these as separate regimes and focus mainly on reference tracking~\cite{hdmi,omniretarget}, creating a precision-flexibility trade-off: dense-tracking policies break down when references are missing or infeasible, while purely goal-conditioned ones lack fine-grained coordination. We seek a unified controller that smoothly transitions between dense plans and sparse intent as information changes.

Despite progress in co-tracking humanoid and object dynamics~\cite{xu2025intermimic}, two bottlenecks hinder unified autonomy. First, kinematic retargeting can yield physically inconsistent demonstrations that fail in contact-rich tasks. Second, existing architectures typically assume a fixed conditioning structure tailored to one input type, and cannot interpret diverse or partial supervision within a consistent framework. Under shifting observability and goals at deployment, this rigidity leads to systemic instability.

To overcome the demonstration bottleneck, we introduce a physics-driven, \textit{neural} retargeting algorithm that transfers large-scale motion capture (MoCap) to humanoid embodiments at scale. Unlike kinematic retargeting~\cite{omniretarget,luo2023perpetual,joao2025gmr}, which struggles to maintain physical consistency in contact-rich tasks, our retargeting is dynamics- and contact-aware by construction. We cast retargeting as simulation-constrained optimization with kinematic, dynamic, and contact constraints, and solve it with reinforcement learning (RL) at scale. Once trained, the policy generates large-scale physically feasible trajectories and generalizes to arbitrary data, enabling augmentation by scaling both objects and motions.

Building on this expanded corpus, we learn a \textit{U}nified mu\textit{LT}imodal cont\textit{R}oller for \textit{A}utonomous humanoid control (ULTRA) that shifts from reference replay to perception-driven, goal-conditioned control. We first train a privileged universal tracker, then distill it into a student that follows diverse goal specifications, from dense references to sparse long-horizon targets (Fig.~\ref{fig:teaser}). This is enabled by (\textbf{i}) unified tokenization with availability masking~\cite{tessler2024maskedmimic}, which keeps a single policy stable when references or modalities are missing; and (\textbf{ii}) a variational skill bottleneck plus RL finetuning~\cite{xu2026interprior} geared toward deployment with realistic perception and sensor noise. The bottleneck resolves ambiguity under sparse goals by maintaining coherent motion, while RL finetuning shifts control from reference-conditioned tracking to closed-loop goal stabilization under partial observability and distribution shift. Together, ULTRA yields one policy that tracks references when available and executes from egocentric perception and sparse intent when they are not.

In summary, ULTRA presents a unified system for practical whole-body loco-manipulation with three components: (\textbf{1}) a physics-driven neural retargeting framework that scales MoCap to humanoid embodiments and supports zero-shot augmentation; (\textbf{ii}) a versatile \textit{multimodal} controller distilled from a privileged tracker that supports reference tracking and goal following across sensing modalities, including blind, MoCap-based, and depth-perception settings; and (\textbf{iii}) simulation and real-world evaluation on Unitree G1, showing that a unified model can outperform tracking baselines when references exist while enabling goal-conditioned behaviors.

\section{Related Work}
\subsection{Motion Retargeting}
Retargeting transfers motion across embodiments with different morphologies. It originated in animation, where inverse-kinematics optimization adapted motions under kinematic constraints~\cite{lee1999hierarchical}, and later evolved into learning-based mappings that amortized transfer for better generalization~\cite{ villegas2018neural}. Humanoid retargeting requires stronger constraints because executability is contact-dependent and subject to joint limits and dynamics. As a result, existing robot retargeting methods trade off efficiency and physical fidelity: kinematic approaches are fast but often under-model dynamics and degrade in contact-rich settings~\cite{park2025demodiffusion,omniretarget,luo2023perpetual,joao2025gmr,tessler2024maskedmimic}, while physics-based retargeting enforces dynamics for physically plausible motions, but relies on expensive optimization, typically per-trajectory RL~\cite{xu2025scalable} or costly sampling-based methods~\cite{pan2025spider}. We target the missing regime: \emph{physics-driven yet scalable} retargeting that preserves interaction semantics without per-trajectory RL. We perform large-scale retargeting with a single unified policy, and enable zero-shot augmentation to expand coverage.

\subsection{Humanoid Whole-Body Locomotion}
Leveraging human motion data to teach humanoid robots complex skills has been widely studied. Early methods often used model-based control (\textit{e.g.}, trajectory optimization and model predictive control) to bridge embodiment and dynamics, while recent learning-based systems achieve precise tracking and agile motion replay~\cite{OmniH2O,exbody2,asap,hover,visualmimic,hdmi,twist,gmt}. Beyond pure tracking, recent work moves toward foundation-style control by distilling large motion corpora into reusable priors, where a single model tracks diverse motions and supports multiple control modes~\cite{unitracker,sonic,beyondmimic,bfm}. Others shape latent priors with adversarial RL~\cite{xue2025leverb,ma2025styleloco,shi2025adversarial,li2025bfm}, but have not shown reliable scaling to large, heterogeneous loco-manipulation corpora.
ULTRA follows the scalable teacher-student distillation paradigm but addresses a key bottleneck: offline distillation is limited by the state coverage of teacher rollouts. While less severe for humanoid-only control with more structured spaces, it becomes acute in high-dimensional robot-object interaction. To address this, we draw inspiration from animation practice~\cite{xu2026interprior} and use a two-stage framework: large-scale distillation followed by RL finetuning that \emph{expands} interaction-state coverage and improves robustness to out-of-distribution (OOD) goals and executions.

\subsection{Humanoid Whole-Body Loco-Manipulation}
Most humanoid motion tracking emphasizes reproducing human motion on the robot and treats environmental dynamics as secondary~\cite{OmniH2O,ben2024homie,sun2025ulc,li2024okami}, which is brittle for contact-rich loco-manipulation. Recent work couples humanoid motion and object interaction via co-tracking and shows strong agility~\cite{omniretarget,hdmi,resmimic,fu2025demohlm}, but often assumes limited data replay or relies on external object state estimation (\textit{e.g.}, motion capture), limiting autonomy under onboard egocentric sensing. Other approaches use hierarchical designs that generate trajectories/keypoints and track them with a universal controller~\cite{DreamControl,visualmimic}; however, stacking a high-level planner on a low-level controller can accumulate error and violate physical constraints. Adversarial motion priors broaden coverage but are typically task-specific, requiring careful objective engineering and scaling poorly to large, heterogeneous loco-manipulation corpora~\cite{wang2025physhsi}. \method~addresses these issues by learning a goal-conditioned policy that unifies dense tracking and sparse task specifications in a \textit{shared} latent space, and by using RL finetuning to induce \textit{closed-loop} behaviors that expand interaction-state coverage. This yields a \emph{versatile} single-policy controller under real-world perception with a \emph{scalable} paradigm that leverages broad motion corpora.

\begin{figure*}
    \centering
    \includegraphics[width=\linewidth]{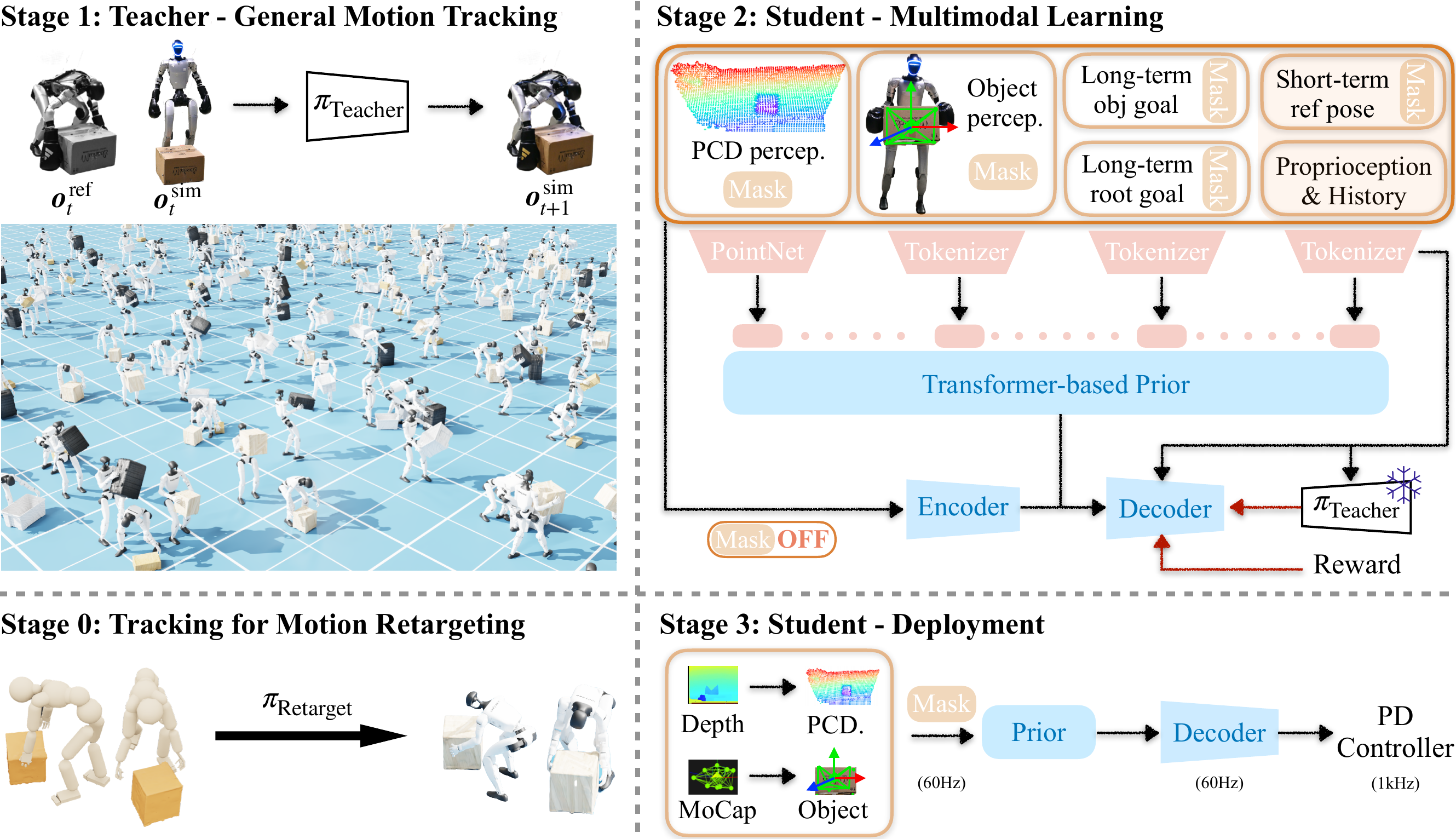}
    \caption{ULTRA follows four stages: (\textbf{i}) \textbf{Neural Retargeting:} an RL policy converts MoCap data into physically feasible G1 rollouts with augmentation; (\textbf{ii}) \textbf{Tracking:} a privileged teacher tracks these rollouts using full state and references; (\textbf{iii}) \textbf{Distillation:} we distill the teacher into a multimodal student for realistic sensing and sparse goals, with additional RL finetuning; (\textbf{iv}) \textbf{Deployment:} the student runs under real sensing, supporting depth input or MoCap-based state estimation.
     \label{fig:method}}
    \vspace{-2em}
\end{figure*}
\section{Problem Formulation and Preliminaries}

\subsection{Task Interface}\label{sec:task-interface}
We study whole-body loco-manipulation tasks where a humanoid interacts with a manipulated object, specified by a \emph{goal} signal $\boldsymbol c \in \mathcal{C}$ that defines the task objective. A rollout succeeds if the terminal outcome satisfies $\boldsymbol c$, \textit{e.g.}, the humanoid root and/or the object reaches target transformations within a tolerance.
At each time step $t$, the policy receives (\textbf{i}) an observation $\boldsymbol o_t \in \mathcal{O}$ and (\textbf{ii}) task conditioning $\boldsymbol c_t$, and outputs an action $\boldsymbol a_t \in \mathcal{A}$. Here $\boldsymbol a_t$ specifies target joint positions executed by a PD controller.

\noindent\textbf{Goal specification.} We consider two forms of $\boldsymbol c$: (\textbf{i}) \emph{dense reference conditioning}, which provides a time-indexed motion reference and thus specifies intermediate motions; and (\textbf{ii}) \emph{sparse goal conditioning}, which specifies long-horizon target transformations for the humanoid root and/or object while leaving intermediate motions underdetermined.

\noindent\textbf{Perception.} Beyond proprioception, we consider two regimes for object sensing: (\textbf{i}) \emph{MoCap-based sensing}, where $\boldsymbol o_t$ includes accurate object pose (\textit{e.g.,} from motion capture); and (\textbf{ii}) \emph{egocentric depth perception}, where $\boldsymbol o_t$ includes an egocentric point cloud from a depth sensor (\textit{e.g.,} head-mounted), from which object state must be inferred.

\subsection{Preliminaries}
Since the controller may rely on partial onboard sensing, we model loco-manipulation as a goal-conditioned \emph{Partially Observable Markov Decision Process} (POMDP). Let $\boldsymbol s_t \in \mathcal{S}$ be the underlying system state (humanoid and scene, including the object), with dynamics $\boldsymbol s_{t+1} \sim \mathcal{T}(\boldsymbol s_t,\boldsymbol a_t)$. The policy acts from $\boldsymbol o_t=\Omega(\boldsymbol s_t)$ and conditioning $\boldsymbol c_t$, producing $\boldsymbol a_t \in \mathcal{A}$. We optimize $\pi(\boldsymbol a_t \mid \boldsymbol o_t,\boldsymbol c_t)$ to maximize expected discounted return:
\(
\max_{\pi}\ \mathbb{E}\big[\sum_{t\ge 0}\gamma^t\,r(\boldsymbol s_t,\boldsymbol a_t,\boldsymbol c_t)\big].
\)
The following sections describe how we use PPO~\cite{ppo} and imitation to learn policies, including the observation/reward design and key techniques for our tasks.

\section{Method}
As shown in Fig.~\ref{fig:method}, \textsc{ULTRA} follows a four-stage training paradigm that couples physics-driven motion retargeting with teacher-student learning. In Stage~1, we learn a \emph{retargeting policy} that maps human MoCap motions to physically feasible humanoid loco-manipulation rollouts. In Stage~2, we train a privileged \emph{teacher policy}, leveraging full state and dense reference trajectories from the retargeted rollouts. In Stage~3, we distill the teacher into a \textit{multimodal student} that operates under perception and sparse goal specifications. Finally, we deploy the student with a separated control mode.

\subsection{General Motion Tracking for Neural Retargeting}\label{sec:retargeting}
Given a human-object demonstration represented by an SMPL-X~\cite{SMPL-X:2019} motion sequence and an object pose trajectory, our goal is to generate a physically feasible rollout on the target humanoid (\textit{e.g.,} Unitree G1) that preserves the overall motion and intended interaction (Fig.~\ref{fig:retarget}). Traditional retargeting solves inverse kinematics under kinematic constraints. We instead cast retargeting as RL-based trajectory optimization: rewards encode tracking, while simulator transitions enforce kinematics, dynamics, and contacts. Following~\cite{xu2025intermimic}, this is well suited for contact-rich loco-manipulation, where contacts are hard to express as kinematic constraints.
As preprocessing, we scale the human-object trajectory to match G1 and define a fixed correspondence from human key links to humanoid counterparts. We then train a \textit{unified} retargeting policy across all motions, producing physically consistent rollouts without per-motion optimization or retraining.
Dense, full-body tracking is brittle under embodiment mismatch and becomes especially fragile during object interaction, where exact link-wise targets may be infeasible and contact often requires deliberate deviations. Our key insight is to combine (\textbf{i}) relaxed tracking that prioritizes end effectors critical for loco-manipulation with (\textbf{ii}) interaction and contact rewards that correct mismatch-induced errors.

\noindent\textbf{Reward.}
We define
\(
r_{\text{track}} = r_p \cdot r_r \cdot r_{\text{obj}} \cdot r_{\text{int}} \cdot r_{\text{ct}} \cdot r_{\text{eng}},
\)
with all terms computed in a heading-aligned humanoid frame. Let $\mathcal{F}$ include only feet and palms. $r_p$ tracks end-effector positions as sparse anchors; $r_r$ matches normalized link directions over a fixed key edge set; and $r_{\text{eng}}$ regularizes joint effort and foot placement. To reduce ambiguity, $r_{\text{obj}}$ tracks object pose/velocities and $r_{\text{int}}$ matches palm-to-surface offsets over sampled object points. We also align contact events by mapping contacts on human links to corresponding humanoid links, yielding $r_{\text{ct}}$.

\noindent\textbf{Observation.}
The policy uses a privileged, reference-aware observation containing simulator state and its deviation from the SMPL-X reference. Since preprocessing establishes a fixed correspondence after scaling/alignment, residuals are well-defined:
\(
\boldsymbol o_t = \big[\boldsymbol o_t^{\text{sim}},\ \boldsymbol o_t^{\text{ref}},\ \boldsymbol o_t^{\Delta}\big].
\)
$\boldsymbol o_t^{\text{sim}}$ includes proprioception and contact signals; $\boldsymbol o_t^{\text{ref}}$ provides selected correspondence-defined reference quantities (including object state); and $\boldsymbol o_t^{\Delta}$ encodes heading-aligned simulation-reference differences. All quantities are expressed in a heading-aligned frame to remove global yaw. 

\noindent\textbf{State initialization and early termination.}
Because we cannot reliably initialize the humanoid from an SMPL-X pose, we do not use reference-state initialization~\cite{peng2018deepmimic}. Each episode starts from a default standing pose, initially tracking the first reference frame to stabilize the humanoid before transitioning to full tracking with smoothly varying weights. We terminate on falls, excessive deviation, or contact mismatch for 20 frames~\cite{xu2025intermimic} to improve sample efficiency.

\noindent\textbf{Simplified actuation.}
Since retargeting is used only to \emph{generate reference rollouts}, we prioritize motion quality and throughput over hardware-faithful control. We use an \textit{idealized} low-level controller in simulation,\textit{e.g.}, with control frequency equal to simulation frequency, enabling stronger, more responsive tracking than onboard PD control. We train without domain randomization or perturbations, and address robustness later in Sec.~\ref{sec:teacher-policy}.

\noindent\textbf{Trajectory and object augmentation.}
RL-based retargeting also enables \emph{flexible augmentation} (Fig.~\ref{fig:aug_comparison}). Since preprocessing already scales positions, we can (\textbf{i}) apply anisotropic scaling along coordinate axes and (\textbf{ii}) scale the manipulated object with different coefficients, while interaction/contact rewards correct imperfections and the simulator enforces physical feasibility. Crucially, these augmentations are handled by a \textit{single retargeting policy without retraining}.

\subsection{Dense Motion Tracking for Teacher Policy}\label{sec:teacher-policy}
Sec.~\ref{sec:retargeting} converts human-object demonstrations into physically feasible G1 rollouts. For downstream imitation, we train a separate privileged teacher $\pi_{\text{teacher}}$ to track these rollouts. The teacher uses the deployment control interface and actuation limits, but trains with privileged state and dense reference residuals to accelerate learning. We randomize physics and inject perturbations to broaden state visitation and teach recovery, producing stable behaviors that provide high-quality supervision for the student.

\noindent\textbf{Observation.}
The teacher uses the same reference-aware observation as retargeting, but does not require cross-embodiment correspondence since the reference is already in the humanoid embodiment.

\noindent\textbf{Dense tracking objective.}
The teacher uses the same reward template, but replaces sparse anchoring with \emph{full} link tracking, together with object, interaction, and contact reward.

\noindent\textbf{Reference initialization and robustness training.}
We initialize from randomly sampled reference frames and include occasional stand still episodes that track standing references, reflecting deployment from a stable standing pose. To improve robustness, we randomize humanoid/object physical properties and inject perturbations, with a short grace period to allow recovery. We use the same early-termination criteria as retargeting and add no observation noise at this stage.

\subsection{Multimodal Student Policy}\label{sec:student-policy}

We distill the privileged teacher into a multimodal student policy $\pi_{\text{student}}$. Unlike the teacher, the student observes only partial state and conditions on whatever modalities are available at test time via an availability mask randomly sampled during training. This retains teacher behavior as a prior while enabling goal-reaching under missing observations.

\noindent\textbf{Multimodal observation with availability mask.}
The student consumes heterogeneous inputs:
\(
\boldsymbol{o}_t^{\text{student}}
=\big[\boldsymbol{o}_t^{\text{proprio}},\ \boldsymbol{o}_t^{\text{goal}},\ \boldsymbol{o}_t^{\text{object}},\ \boldsymbol{o}_t^{\text{pcd}},\ \boldsymbol{m}_t\big].
\)
$\boldsymbol{o}_t^{\text{proprio}}$ contains proprioception (\textit{e.g.,} joint states, IMU), $\boldsymbol{o}_t^{\text{object}}$ provides object state (\textit{e.g.,} MoCap), and $\boldsymbol{o}_t^{\text{pcd}}$ is an egocentric point cloud (\textit{e.g.,} egocentric camera). $\boldsymbol{o}_t^{\text{goal}}$ encodes task objectives and commands, including (\textbf{i}) long-horizon object transforms, (\textbf{ii}) long-horizon humanoid root transforms, and (\textbf{iii}) next-frame humanoid local state changes for tracking. We also include discretized commands (\textit{e.g}., stand still) for deployment. $\boldsymbol{m}_t$ indicates which modalities are present.

\noindent\textbf{Distillation.}
We collect data with a DAgger-style loop~\cite{ross2011reduction}: we roll out with the teacher initially, gradually shift to the student, and query the teacher on visited states to obtain $\boldsymbol{a}_t^{\text{teacher}}$. During training, an encoder
$q_\phi(\boldsymbol{z}_t^{\mathrm{res}} \mid \boldsymbol{o}_t^{\text{student}}, \boldsymbol{o}_t^{\text{teacher}})$
infers a latent residual~\cite{tessler2024maskedmimic} using privileged teacher inputs, while a prior
$p_\theta(\boldsymbol{z}_t^{\mathrm{prior}} \mid m(\boldsymbol{o}_t^{\text{student}}))$
predicts a latent from masked student observations ($m(\cdot)$ applies $\boldsymbol{m}_t$). We combine them as
${\boldsymbol{z}}_t=\boldsymbol{z}_t^{\mathrm{prior}}+\boldsymbol{z}_t^{\mathrm{res}}$ and sample actions
$\boldsymbol{a}_t^{\text{student}} \sim \pi_{\text{student}}(\boldsymbol{a}_t \mid \boldsymbol{o}_t^{\text{student}}, {\boldsymbol{z}}_t)$.
We implement $\pi_{\text{student}}$ with a transformer-based encoder~\cite{transformer} that projects each modality into shared tokens; $\boldsymbol{m}_t$ gates tokens and modulates cross-modal attention to ignore missing inputs. At deployment, we sample $\boldsymbol{z}_t$ from the prior only.

\noindent\textbf{Training objective.}
We match teacher actions while aligning the prior with the privileged posterior:
\begin{equation}
\begin{aligned}
\mathcal{L}
&={\|\boldsymbol{a}_t^{\text{student}}-\boldsymbol{a}_t^{\text{teacher}}\|_2^2}
+\mathcal{L}_{\text{aux}}
\\&+\lambda_{\text{KL}}{D_{\text{KL}}\!\left(q_\phi(\boldsymbol{z}_t\mid \boldsymbol{o}_t^{\text{student}},\boldsymbol{o}_t^{\text{teacher}})
\ \|\ p_\theta(\boldsymbol{z}_t\mid \boldsymbol{o}_t^{\text{student}})\right)}.
\end{aligned}
\end{equation}
$\mathcal{L}_{\text{aux}}$ uses reconstruction heads (recovering masked modalities) to encourage $\boldsymbol{z}_t$ to retain task-relevant information.

\noindent\textbf{Curriculum learning.}
Beyond DAgger, we use two curricula to keep the prior effective under partial observability: we progressively increase modality-masking probability, and anneal $\lambda_{\text{KL}}$ and auxiliary weights to avoid posterior collapse while preserving latent skill diversity.

\noindent\textbf{Shortcut for tracking.}
For local-goal tracking, behavior is largely deterministic, so a stochastic latent helps less. We add a residual shortcut (with the mask) from the full-body goal directly to the decoder, preserving low-level reference information and stabilizing decoding (Fig.~\ref{fig:method}).

\noindent\textbf{RL finetuning.}
We perform RL finetuning on top of the distilled student by switching a subset of parallel environments to a goal-reaching objective while continuing distillation updates. Following~\cite{xu2026interprior}, we partition simulators into (\textbf{i}) distillation environments replaying reference motions with imitation losses, and (\textbf{ii}) RL environments optimizing task success under state/goal perturbations. We sample random offsets for the object goal, humanoid root goal, and their initializations\rebuttal{, and inject simulated point-cloud noise (jitter, dropout, occlusion) for alleviating sim-to-real gap}.

\noindent\textbf{Deployment versatility.}
At test time, the student receives only $\boldsymbol{o}_t^{\text{student}}$ and samples $\boldsymbol{z}_t$ from the prior. With the same parameters, modality masking enables: (\textbf{i}) high-fidelity tracking by unmasking local reference (Fig.~\ref{fig:teaser} \textbf{Top}), (\textbf{ii}) goal-conditioned control by masking local reference and unmasking long-horizon goals (Fig.~\ref{fig:teaser} \textbf{Middle}), and (\textbf{iii}) vision-based manipulation by masking MoCap object state while unmasking point clouds (Fig.~\ref{fig:teaser} \textbf{Bottom}).

\begin{figure}
    \centering
    \includegraphics[width=\linewidth]{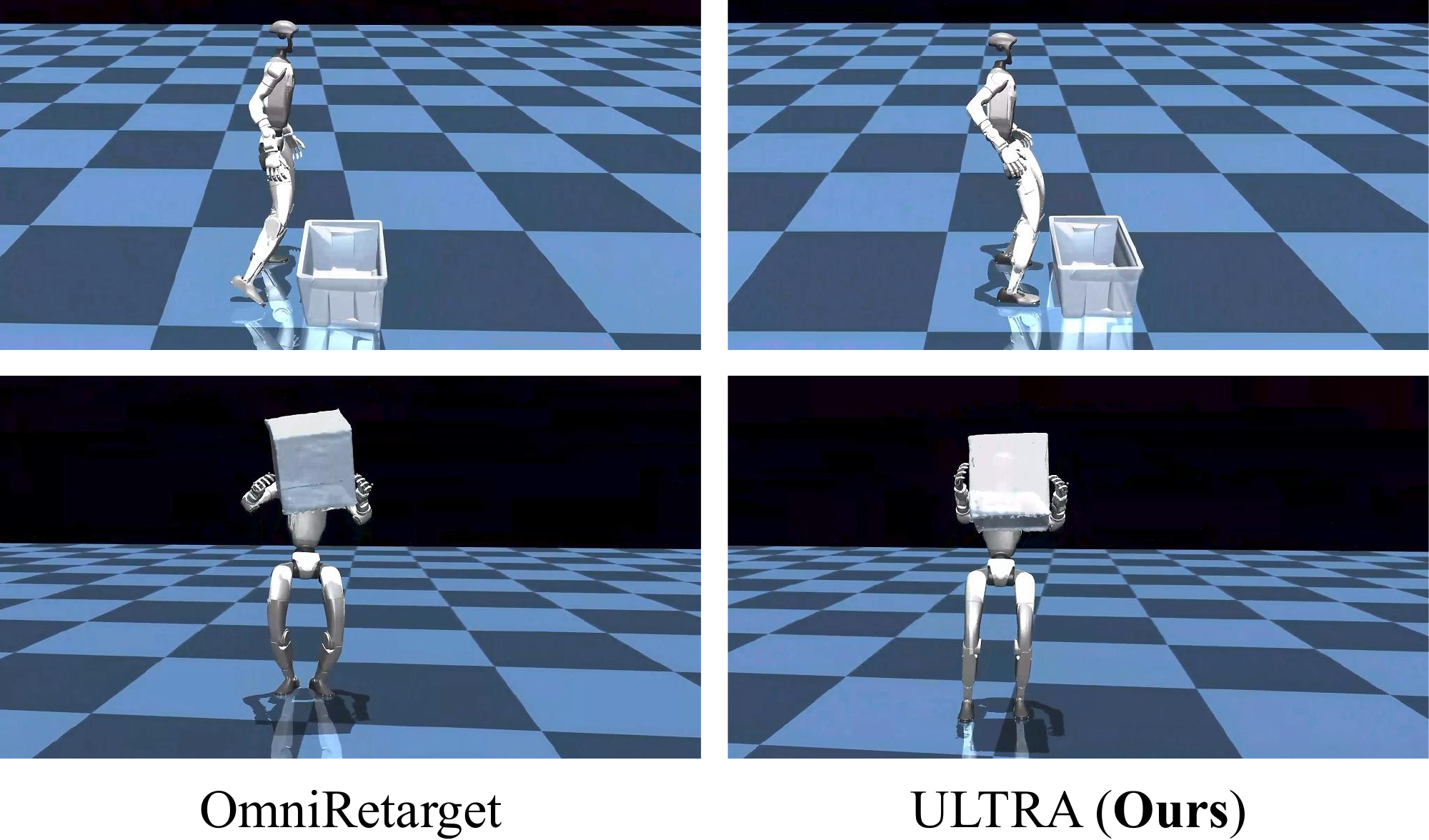}
    \caption{Qualitative comparison of our retargeting and OmniRetarget~\cite{omniretarget} at the same frame/sequence. \textbf{Top:} final frame; the baseline shows undesired standing foot placement. \textbf{Bottom:} a contact frame; ours yields more stable contacts.}
    \label{fig:retarget}
    \vspace{-2em}
\end{figure}

\begin{table*}
\caption{Motion-tracking evaluation in IsaacGym. All methods are trained/evaluated on our data unless noted. \cellcolor{green!12}\textbf{Green} highlights our primary tracking controller.}
\label{tab:tracking_main}
\centering
\resizebox{\textwidth}{!}{%
\begingroup
\setlength{\tabcolsep}{2pt}
\renewcommand{\arraystretch}{1.15}

\newcommand{\bestours}[1]{\cellcolor{green!12}\textbf{#1}}
\newcommand{\upperbound}[1]{\cellcolor{gray!4}\textit{#1}}

\begin{tabular}{l cc ccccc cc ccccc}
\toprule
& \multicolumn{7}{c}{\cellcolor{blue!8}\textbf{In-Distribution (ID)}} & \multicolumn{7}{c}{\cellcolor{orange!8}\textbf{Out-of-Distribution (OOD)}} \\
\cmidrule(lr){2-8} \cmidrule(lr){9-15}
& $\text{Succ}\uparrow$ & $\text{Succ}\uparrow$ & \multicolumn{3}{c}{Humanoid} & \multicolumn{2}{c}{Object}
& $\text{Succ}\uparrow$ & $\text{Succ}\uparrow$ & \multicolumn{3}{c}{Humanoid} & \multicolumn{2}{c}{Object} \\
\cmidrule(lr){4-6} \cmidrule(lr){7-8} \cmidrule(lr){11-13} \cmidrule(lr){14-15}
\textbf{Method} & (Humanoid)
 & (+Object)
 &
$E_\text{g-mpjpe}\downarrow$ & $E_\text{mpjpe}\downarrow$ & $E_\text{jitter}\downarrow$ &
$E_\text{pos}\downarrow$ & $E_\text{rot}\downarrow$ & (Humanoid)
 & (+Object)
 &
$E_\text{g-mpjpe}\downarrow$ & $E_\text{mpjpe}\downarrow$ & $E_\text{jitter}\downarrow$ &
$E_\text{pos}\downarrow$ & $E_\text{rot}\downarrow$ \\
\midrule

\rowcolor{gray!15}
\multicolumn{15}{c}{\textbf{(a) Unified Multimodal Controller}} \\
{\method{}} (\textbf{Ours}) &
67.30{\scriptsize$\pm$0.12} &
57.44{\scriptsize$\pm$0.40} &
13.49{\scriptsize$\pm$0.14} &
\phantom{0}5.89{\scriptsize$\pm$0.02} &
\phantom{0}6.27{\scriptsize$\pm$0.00} &
53.42{\scriptsize$\pm$0.21} &
65.44{\scriptsize$\pm$0.37} &
70.57{\scriptsize$\pm$0.54} &
52.00{\scriptsize$\pm$0.44} &
35.55{\scriptsize$\pm$0.23} &
14.67{\scriptsize$\pm$0.08} &
\phantom{0}6.81{\scriptsize$\pm$0.01} &
56.52{\scriptsize$\pm$0.35} &
67.60{\scriptsize$\pm$0.58} \\
\midrule

\rowcolor{gray!15}
\multicolumn{15}{c}{\textbf{(b) Privileged Teacher}} \\
\rowcolor{gray!8}
\upperbound{ULTRA Teacher} &
\upperbound{97.57{\scriptsize$\pm$0.05}} &
\upperbound{89.79{\scriptsize$\pm$0.11}} &
\upperbound{12.98{\scriptsize$\pm$0.30}} &
\upperbound{\phantom{0}5.64{\scriptsize$\pm$0.05}} &
\upperbound{14.81{\scriptsize$\pm$0.08}} &
\upperbound{17.15{\scriptsize$\pm$0.03}} &
\upperbound{23.28{\scriptsize$\pm$0.33}} &
\upperbound{97.12{\scriptsize$\pm$0.43}} &
\upperbound{81.33{\scriptsize$\pm$0.78}} &
\upperbound{19.14{\scriptsize$\pm$0.48}} &
\upperbound{\phantom{0}7.94{\scriptsize$\pm$0.11}} &
\upperbound{15.91{\scriptsize$\pm$0.08}} &
\upperbound{25.57{\scriptsize$\pm$0.28}} &
\upperbound{33.49{\scriptsize$\pm$0.37}} \\
\midrule

\rowcolor{gray!15}
\multicolumn{15}{c}{\textbf{(c) General Motion Tracking}} \\
\method{} (RL) &
54.47{\scriptsize$\pm$0.43} &
41.78{\scriptsize$\pm$0.31} &
49.30{\scriptsize$\pm$0.32} &
16.23{\scriptsize$\pm$0.11} &
20.04{\scriptsize$\pm$0.15} &
47.48{\scriptsize$\pm$0.31} &
60.53{\scriptsize$\pm$0.09} &
53.38{\scriptsize$\pm$0.98} &
23.54{\scriptsize$\pm$0.24} &
68.11{\scriptsize$\pm$0.26} &
22.22{\scriptsize$\pm$0.01} &
17.46{\scriptsize$\pm$0.09} &
66.17{\scriptsize$\pm$0.54} &
59.44{\scriptsize$\pm$0.73} \\
{\method{}} (Distillation) &
\bestours{85.03{\scriptsize$\pm$3.00}} &
\bestours{77.15{\scriptsize$\pm$0.57}} &
\bestours{15.45{\scriptsize$\pm$0.08}} &
\bestours{\phantom{0}6.84{\scriptsize$\pm$0.04}} &
\bestours{\phantom{0}8.12{\scriptsize$\pm$0.01}} &
\bestours{25.48{\scriptsize$\pm$0.48}} &
\bestours{33.97{\scriptsize$\pm$0.58}} &
\bestours{86.63{\scriptsize$\pm$0.50}} &
\bestours{52.74{\scriptsize$\pm$0.04}} &
\bestours{35.01{\scriptsize$\pm$0.31}} &
\bestours{13.48{\scriptsize$\pm$0.10}} &
\bestours{\phantom{0}9.35{\scriptsize$\pm$0.01}} &
\bestours{36.18{\scriptsize$\pm$0.30}} &
\bestours{38.18{\scriptsize$\pm$0.29}} \\
\midrule

HDMI~\cite{hdmi} &
13.07{\scriptsize$\pm$0.20} &
\phantom{0}9.94{\scriptsize$\pm$0.38} &
92.77{\scriptsize$\pm$0.56} &
26.90{\scriptsize$\pm$0.10} &
26.13{\scriptsize$\pm$0.60} &
78.93{\scriptsize$\pm$0.42} &
70.23{\scriptsize$\pm$0.62} &
13.92{\scriptsize$\pm$0.78} &
12.95{\scriptsize$\pm$0.30} &
87.07{\scriptsize$\pm$0.44} &
27.54{\scriptsize$\pm$0.06} &
29.19{\scriptsize$\pm$0.38} &
77.33{\scriptsize$\pm$2.27} &
71.16{\scriptsize$\pm$0.48} \\
OmniRetarget$^\dagger$~\cite{omniretarget} &
41.27{\scriptsize$\pm$1.17} &
21.90{\scriptsize$\pm$0.29} &
62.96{\scriptsize$\pm$1.43} &
15.37{\scriptsize$\pm$0.17} &
39.35{\scriptsize$\pm$0.57} &
77.94{\scriptsize$\pm$3.52} &
66.47{\scriptsize$\pm$1.15} &
33.36{\scriptsize$\pm$0.39} &
20.78{\scriptsize$\pm$0.13} &
74.80{\scriptsize$\pm$0.34} &
16.23{\scriptsize$\pm$0.15} &
49.52{\scriptsize$\pm$0.52} &
55.11{\scriptsize$\pm$2.32} &
62.44{\scriptsize$\pm$0.77} \\
OmniRetarget~\cite{omniretarget} &
51.34{\scriptsize$\pm$0.67} &
20.91{\scriptsize$\pm$0.52} &
67.12{\scriptsize$\pm$0.86} &
\phantom{0}7.43{\scriptsize$\pm$0.07} &
39.92{\scriptsize$\pm$1.44} &
60.67{\scriptsize$\pm$0.54} &
67.03{\scriptsize$\pm$0.19} &
46.71{\scriptsize$\pm$0.74} &
25.82{\scriptsize$\pm$0.52} &
68.34{\scriptsize$\pm$0.82} &
\phantom{0}8.98{\scriptsize$\pm$0.19} &
40.08{\scriptsize$\pm$1.77} &
58.57{\scriptsize$\pm$2.37} &
66.70{\scriptsize$\pm$0.84} \\
\bottomrule
\multicolumn{15}{l}{\footnotesize $^\dagger$ Trained/evaluated on original OmniRetarget dataset.} \\
\end{tabular}
\endgroup}
\vspace{-2em}
\end{table*}

\section{Experimental Results}
We evaluate \method{} end-to-end for autonomous whole-body loco-manipulation, from data generation to real-world transfer. We ask: (\textbf{i}) Can we retarget human-object MoCap into physically consistent rollouts with stable contacts and minimal sliding/penetration? (\textbf{ii}) Under dense references, can the student match a privileged teacher and specialized trackers? (\textbf{iii}) Under sparse goals, does RL finetuning improve robustness and yield a semantically organized latent skill space? (\textbf{iv}) Can one policy transfer to a real humanoid without test-time references? We evaluate four axes: retargeting, tracking, goal execution, and real deployment on Unitree G1.

\subsection{Experimental Setup}
\noindent\textbf{Simulation.} We train in IsaacGym~\cite{makoviychuk2021isaac} with GPU-parallel environments and validate key results in MuJoCo~\cite{todorov2012mujoco}. Real trials use a physical Unitree G1~\cite{unitreeg1}.

\begin{figure}
    \centering
    \includegraphics[width=\linewidth]{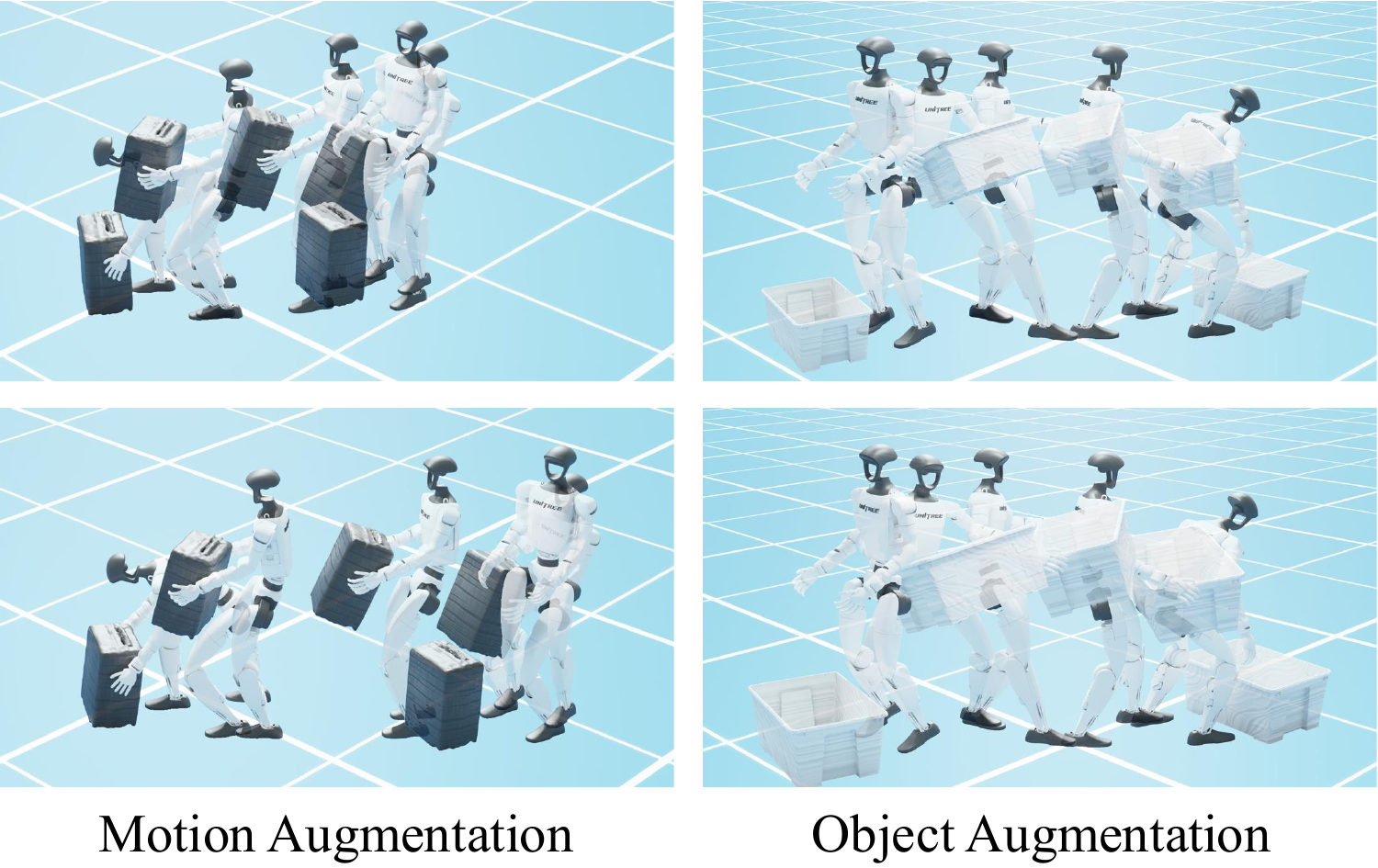}
    \caption{Zero-shot augmentation with the retargeting policy. \textbf{Left:} trajectory scaling. \textbf{Right:} object scaling. Motions remain plausible, enabling scalable data augmentation.}
    \label{fig:aug_comparison}
    \vspace{-2em}
\end{figure}

\noindent\textbf{Dataset.}
We use OMOMO~\cite{li2023object} human-object MoCap, with the corrected subset from~\cite{xu2025intermimic} for a fair comparison with~\cite{omniretarget}. \rebuttal{We focus on four box-like objects compatible with rigid hands.} We retarget all sequences with our RL-based strategy (Sec.~\ref{sec:retargeting}) and augment via anisotropic trajectory scaling and object resizing, yielding a $\sim6\times$ larger corpus (Fig.~\ref{fig:aug_comparison}). We adopt the same train/test split for \textbf{ID} evaluation and define \textbf{OOD} by held-out motions and novel object scales.

\subsection{Motion Retargeting}\label{sec:exp-retarget}
\noindent\textbf{Baselines.}
We compare against: (\textbf{i}) PHC~\cite{luo2023perpetual} (human-only), (\textbf{ii}) GMR~\cite{joao2025gmr} (human-only), and (\textbf{iii}) OmniRetarget~\cite{omniretarget} (human-object interaction). All retarget from the same OMOMO subset processed by~\cite{xu2025intermimic}.

\noindent\textbf{Metrics.}
We measure: (\textbf{i}) \emph{penetration} (duration, max depth) between humanoid/object/environment~\cite{omniretarget}; (\textbf{ii}) \emph{foot skating} (sliding duration, max tangential stance velocity), with stance defined geometrically (foot within 2\,cm of ground) to avoid noisy MoCap stance labels; and (\textbf{iii}) \emph{contact floating}, the duration of lost hand-object contact during transport, detected via MuJoCo contact queries.

\noindent\textbf{Quantitative evaluation.}
Table~\ref{tab:interaction_quality} shows ULTRA outperforms baselines across nearly all metrics/categories: lowest foot-skating duration/velocity and much less contact floating (near-zero on Largebox/Suitcase), while also reducing penetration. We attribute this to physics-aware retargeting that enforces contact/dynamics, keeping stance feet planted and preserving hand-object contact when lifting.

\noindent\textbf{Qualitative evaluation.}
Fig.~\ref{fig:retarget} shows more accurate hand/foot placement than OmniRetarget, whose kinematic formulation often breaks contact consistency and yields unnatural configurations relative to the object and ground.

\noindent\textbf{Effectiveness of data augmentation.}
Our augmentation diversifies motions without retraining the retargeter and applies along the full trajectory (not only the initial frame), producing temporally consistent variations (Fig.~\ref{fig:aug_comparison}). This improves downstream generalization: in Table~\ref{tab:tracking_main}, OmniRetarget retrained on our augmented data attains substantially higher OOD tracking success than when trained on its original dataset, confirming broader state/skill coverage.

\begin{table*}[t]
\centering
\setlength{\tabcolsep}{6pt}
\renewcommand{\arraystretch}{1.15}

\begin{minipage}[t]{0.55\linewidth}
\centering
\caption{Physical interaction quality for retargeting. Ours is better.}
\label{tab:interaction_quality}

\resizebox{\linewidth}{!}{%
\begingroup
\setlength{\tabcolsep}{6pt}
\renewcommand{\arraystretch}{1.15}
\begin{tabular}{l cc cc c}
\toprule
\textbf{Method}
& \multicolumn{2}{c}{\textbf{Penetration}}
& \multicolumn{2}{c}{\textbf{Foot Skating}}
& \textbf{Contact Floating} \\
\cmidrule(lr){2-3}
\cmidrule(lr){4-5}
\cmidrule(lr){6-6}
& Duration $\downarrow$
& Max Depth (cm) $\downarrow$
& Duration $\downarrow$
& Max Vel. (cm/s) $\downarrow$
& Duration $\downarrow$ \\
\midrule
\rowcolor{gray!15}
\multicolumn{6}{l}{\textbf{Largebox}} \\
PHC~\cite{luo2023perpetual}  & 0.908 $\pm$ 0.125 & 0.073 $\pm$ 0.048 & 0.303 $\pm$ 0.145 & 0.032 $\pm$ 0.022 & 0.025 $\pm$ 0.054 \\
GMR~\cite{joao2025gmr}  & 0.522 $\pm$ 0.259 & 0.086 $\pm$ 0.053 & 0.366 $\pm$ 0.317 & 0.029 $\pm$ 0.020 & 0.111 $\pm$ 0.171 \\
OmniRetarget~\cite{omniretarget} & \textbf{0.000 $\pm$ 0.002} & 0.013 $\pm$ 0.002 & 0.205 $\pm$ 0.106 & 0.035 $\pm$ 0.019 & 0.231 $\pm$ 0.224 \\
ULTRA (\textbf{Ours})
& 0.008 $\pm$ 0.030
& \textbf{0.012 $\pm$ 0.002}
& \textbf{0.061 $\pm$ 0.031}
& \textbf{0.018 $\pm$ 0.010}
& \textbf{0.015 $\pm$ 0.063} \\
\midrule
\rowcolor{gray!15}
\multicolumn{6}{l}{\textbf{Suitcase}} \\
PHC~\cite{luo2023perpetual}  & 0.914 $\pm$ 0.119 & 0.077 $\pm$ 0.051 & 0.286 $\pm$ 0.147 & 0.035 $\pm$ 0.024 & 0.032 $\pm$ 0.065 \\
GMR~\cite{joao2025gmr}  & 0.571 $\pm$ 0.265 & 0.105 $\pm$ 0.050 & 0.399 $\pm$ 0.368 & 0.028 $\pm$ 0.018 & 0.142 $\pm$ 0.175 \\
OmniRetarget~\cite{omniretarget} & 0.003 $\pm$ 0.016 & \textbf{0.012 $\pm$ 0.002} & 0.264 $\pm$ 0.141 & 0.040 $\pm$ 0.021 & 0.404 $\pm$ 0.279 \\
ULTRA (\textbf{Ours})
& \textbf{0.002 $\pm$ 0.013}
& 0.017 $\pm$ 0.019
& \textbf{0.062 $\pm$ 0.045}
& \textbf{0.017 $\pm$ 0.008}
& \textbf{0.008 $\pm$ 0.040} \\
\bottomrule
\end{tabular}
\endgroup}
\vspace{-8mm}
\end{minipage}
\hfill
\begin{minipage}[t]{0.43\linewidth}
\centering
\vspace{1pt}

\begin{subfigure}[t]{0.48\linewidth}
\centering
\includegraphics[width=\linewidth]{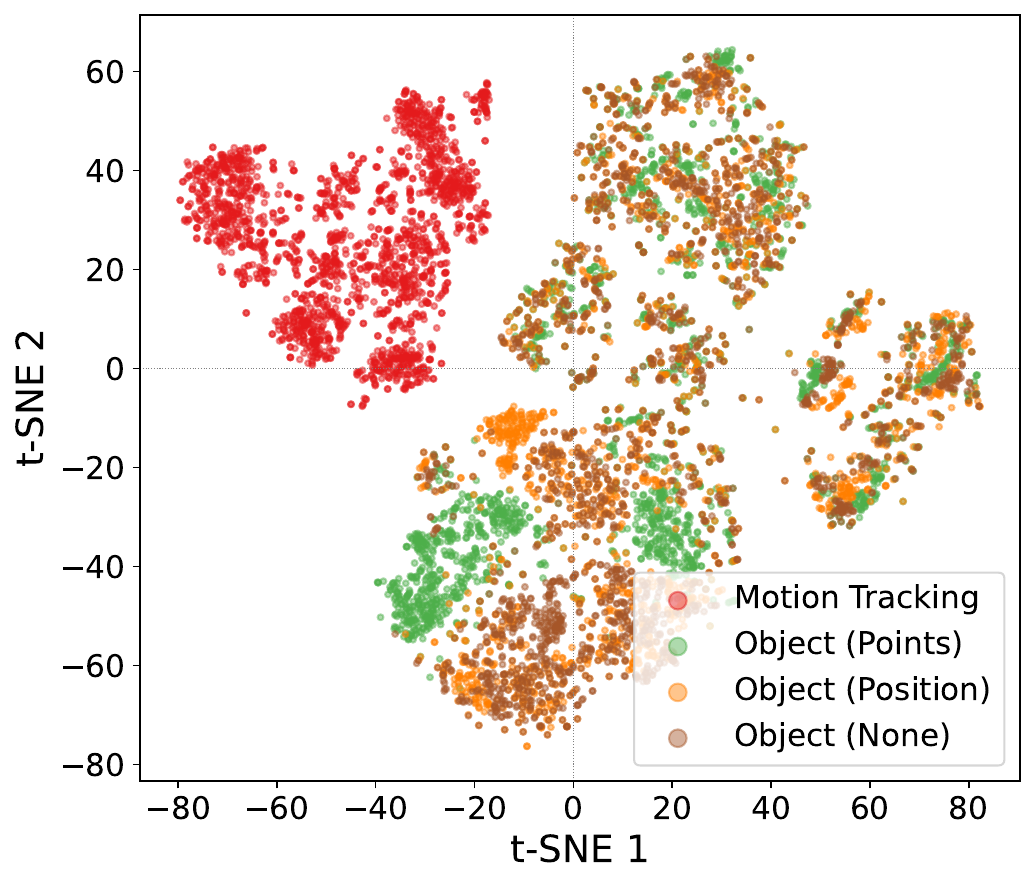}
\end{subfigure}\hfill
\begin{subfigure}[t]{0.48\linewidth}
\centering
\includegraphics[width=\linewidth]{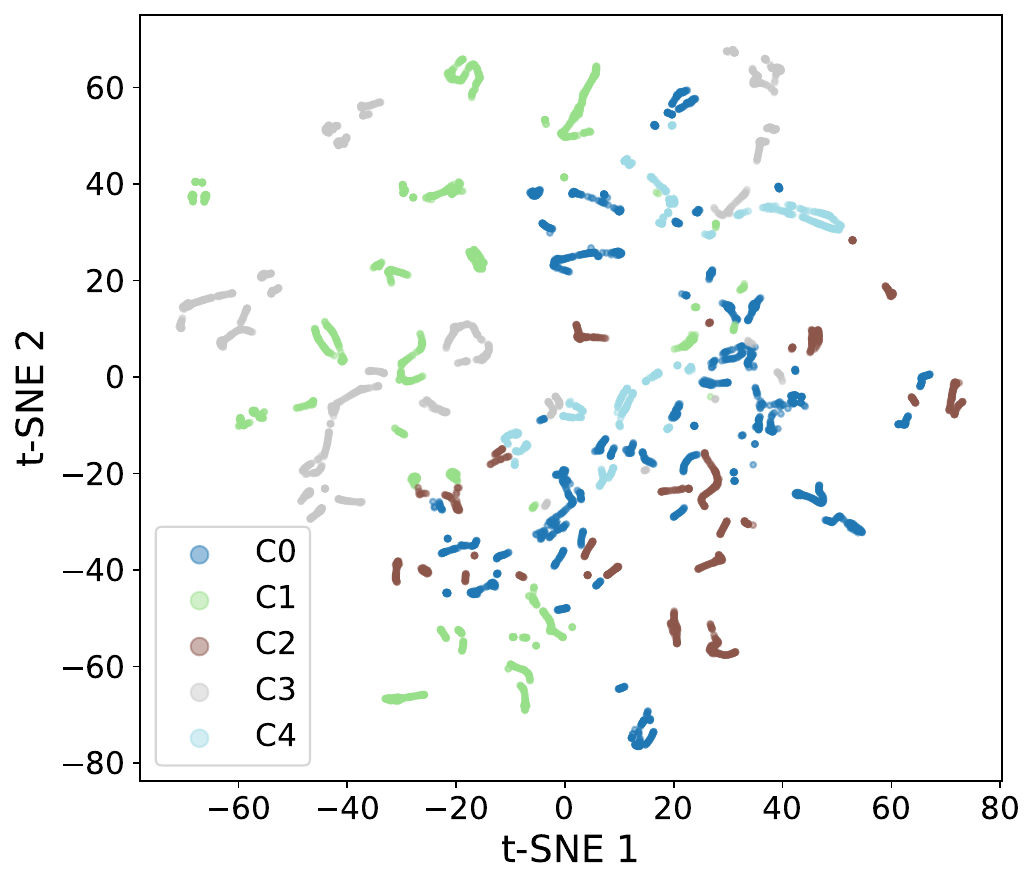}
\end{subfigure}

\vspace{-8pt}
\captionof{figure}{\textbf{Left:} embeddings under different modalities; aside from tracking, embeddings largely mix, indicating a shared latent space. \textbf{Right:} latents cluster by text labels (C0--C4), showing semantic structure.}
\label{fig:delta_tsne}
\end{minipage}
\vspace{-2em}
\end{table*}

\subsection{General Motion Tracking}\label{sec:exp-tracking}
\noindent\textbf{Baselines.}
We evaluate dense tracking (full reference provided) against: (\textbf{i}) OmniRetarget$^\dagger$ (original data), (\textbf{ii}) OmniRetarget retrained on our augmented set, and (\textbf{iii}) HDMI~\cite{hdmi} adapted to our setting. We also report ULTRA ablations: (\textbf{i}) direct RL under student observations (tracking only), (\textbf{ii}) tracking-only distillation, and (\textbf{iii}) all-task unified training. The privileged teacher is an upper bound.

\noindent\textbf{Metrics.}
We report success (Succ): no fall and per-frame $E_{\text{g-mpjpe}}<0.3$\,m and $E_{\text{pos}}<0.3$\,m; we also report humanoid-only success. Tracking errors include $E_{\text{g-mpjpe}}$, $E_{\text{mpjpe}}$, $E_{\text{jitter}}$, and object errors $E_{\text{pos}}$, $E_{\text{rot}}$.

\noindent\textbf{Results.}
Table~\ref{tab:tracking_main} shows ULTRA strongly outperforms baselines for humanoid-object tracking, especially under OOD motions/object scales. HDMI often becomes unstable at our scale and fails to converge. OmniRetarget trains smoothly but frequently fails manipulation: humanoid success is reasonable, but drops when object tracking is required, likely due to missing explicit object observations and a default-to-locomotion failure mode. ULTRA closes this gap via a privileged teacher with object/contact signals and distillation that preserves closed-loop tracking under partial observability. \rebuttal{The residual teacher-student gap (97.57\% vs.\ 67.30\% ID) reflects this unification rather than under-fitting, as a tracking-only student already reaches 85.03\%.}

\noindent\textbf{Distillation vs.\ direct RL.}
Direct RL under partial observations must learn stabilization and sustained contact jointly, so early failures dominate and training can collapse. Distillation transfers contact-aware corrections learned with privileged state and dense references, yielding higher success.

\begin{figure}
    \centering
    \includegraphics[width=\linewidth]{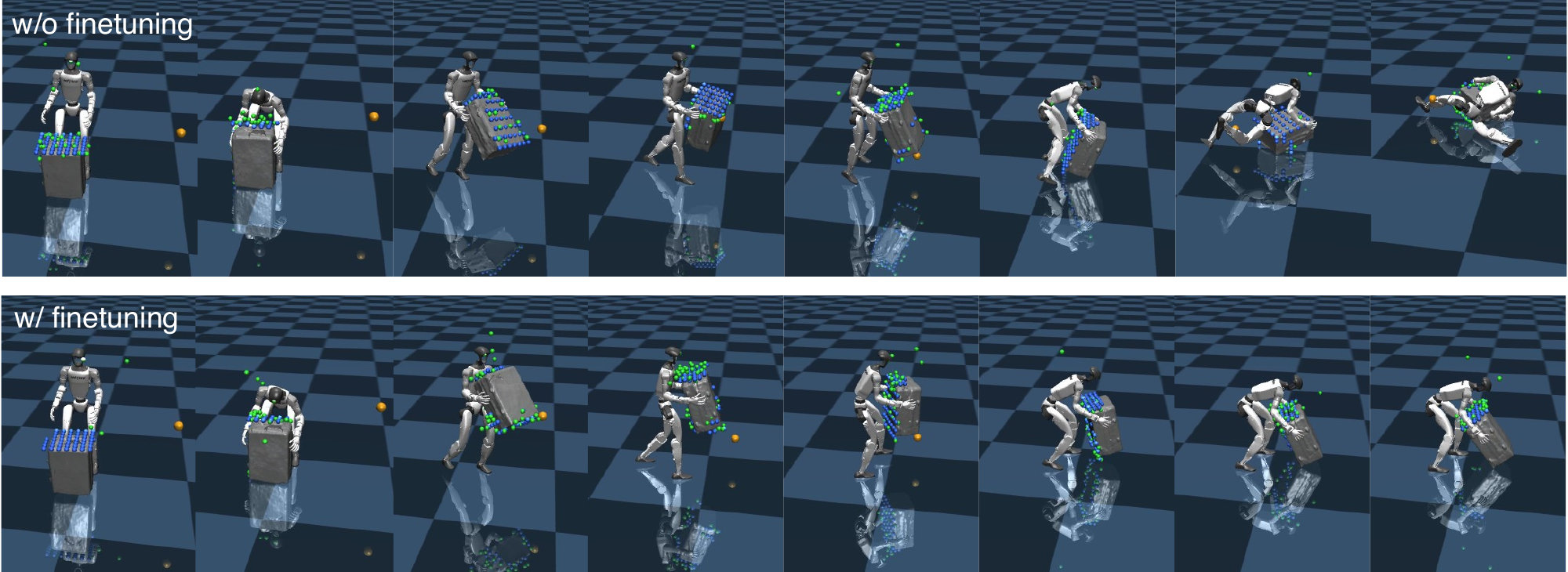}
    \caption{Sim-to-sim comparison for egocentric goal following. Blue/green: point cloud without/with noise; yellow: object goal. \textbf{Top:} without finetuning. \textbf{Bottom:} with finetuning.}
    \label{fig:mujoco}
    \vspace{-2em}
\end{figure}

\subsection{Goal-Conditioned Following}\label{sec:exp-goal}
\noindent\textbf{Metric.}
Success (Succ): no fall and terminal state within $0.3$\,m of the goal.

\noindent\textbf{Comparisons.}
Tracking-only baselines (\textit{e.g.,} HDMI~\cite{hdmi}, OmniRetarget~\cite{omniretarget}) require dense references and are inapplicable; we compare ULTRA to ablations.

\noindent\textbf{Tasks.}
We test two perception regimes: (\textbf{i}) \textit{MoCap} object pose and (\textbf{ii}) \textit{egocentric vision} from a noisy depth point cloud. In both, we provide only long-horizon object target.

\noindent\textbf{RL finetuning expands OOD coverage.}
Table~\ref{tab:ultra_succ} and Fig.~\ref{fig:mujoco} show finetuning yields modest ID gains but large OOD gains under random goal offsets (nearly doubling under point clouds and tripling under position-only). This suggests finetuning expands interaction-state coverage and reinforces closed-loop recovery beyond the demonstration manifold.

\noindent\textbf{Latent structure.}
Fig.~\ref{fig:delta_tsne} shows the latent separates tracking vs.\ goal following across modalities while remaining on a shared manifold. Tracking stays distinct because we bypass the stochastic latent via a residual shortcut when local tracking goals are provided (Sec.~\ref{sec:student-policy}), leaving the latent to capture multimodality under sparse goals. The latent also aligns with text-derived semantic clusters (MiniLM~\cite{wang2020minilm} + K-Means), indicating organization by control regime and intent.

\subsection{Real-World Deployment}\label{sec:exp-real}
\noindent\textbf{Tasks.}
We deploy \method{} on a physical Unitree G1. The student runs onboard at the control frequency with proprioception and, when available, OptiTrack object pose. For dense tracking, we test OMOMO subsets (bimanual box lift/carry, suitcase transport) with household objects. For goal-conditioned control, we provide no motion references and specify future object transforms via keyboard commands. \rebuttal{Beyond picking, our real-robot evaluation also covers locomotion with other contact pattern, including kicking boxes.}

\noindent\textbf{Point cloud extraction.}
For egocentric perception, we extract object point clouds from depth only: back-project depth pixels using calibrated intrinsics, crop a forward ROI, remove the ground plane, take the dominant cluster as the box, and downsample to a fixed size for policy input.

\noindent\textbf{Quantitative evaluation.}
Table~\ref{tab:real_world} reports success rates: the policy reliably grasps/transports on hardware and achieves reasonable sparse-goal success under out-of-distribution operator commands, including composed motions.

\noindent\textbf{Failure analysis.}
Failures mainly arise from (\textbf{i}) friction gaps causing occasional grasp slip, (\textbf{ii}) depth noise/occlusion breaking point-cloud extraction, and (\textbf{iii}) disturbances beyond the recovery margin learned with domain randomization, motivating future tactile integration.

\begin{table}
\caption{Sim-to-sim success rate on Mujoco across goal types with in-distributional (ID) goals from training and out-of-distribution (OOD) goals with random offsets, and across perception with egocentric point clouds or object position with no shape. Policies are trained in IsaacGym and evaluated in MuJoCo with 20 selected motion per setting.}
\label{tab:ultra_succ}
\centering
\setlength{\tabcolsep}{6pt}
\renewcommand{\arraystretch}{1.15}
\resizebox{0.9\columnwidth}{!}{%
\begin{tabular}{c c c c c}
\toprule
\multirow{2}{*}{\textbf{RL finetuning}} &
\multicolumn{2}{c}{\textbf{ID Goals}} &
\multicolumn{2}{c}{\cellcolor{gray!10}\textbf{OOD Goals}} \\
\cmidrule(lr){2-3} \cmidrule(lr){4-5}
& \textbf{Points} & \textbf{Position} &
\cellcolor{gray!10}\textbf{Points} & \cellcolor{gray!10}\textbf{Position} \\
\midrule
\xmark & 16 / 20 & 14 / 20 &
\cellcolor{gray!10}5 / 20 & \cellcolor{gray!10}4 / 20 \\
\cmark & 19 / 20 & 16 / 20 &
\cellcolor{gray!10}\textbf{9 / 20} & \cellcolor{gray!10}\textbf{12 / 20} \\
\rowcolor{gray!05}
$\Delta$ (RL gain) &
+18.8\% & +14.3\% &
\cellcolor{gray!10}\textbf{+80.0\%} &
\cellcolor{gray!10}\textbf{+200.0\%} \\
\bottomrule
\end{tabular}%
}
\vspace{-1em}
\end{table}

\begin{table}[t]
\caption{
Real-world success rates on the OMOMO subset using a Unitree G1 humanoid.
Each task is evaluated over two trials.
MoCap provides object pose tracking for non-egocentric control modes, while the egocentric
setting relies only on onboard sensing.
MoCap is used for success evaluation in all settings.
Dense reference tracking is direction-agnostic and thus reported as a single success rate.
}
\label{tab:real_world}
\centering
\setlength{\tabcolsep}{6pt}
\renewcommand{\arraystretch}{1.15}
\begin{tabular}{lcc}
\toprule
\textbf{Setting} & \textbf{Vertical} & \textbf{Lateral} \\
\midrule
Dense Reference Tracking
& \multicolumn{2}{c}{\textbf{73\%} (19/26)} \\

Sparse Goal Following (MoCap)
& \textbf{80\%} (8/10)
& \textbf{90\%} (9/10) \\

Sparse Goal Following (Egocentric)
& \textbf{50\%} (5/10)
& \textbf{60\%} (6/10) \\
\bottomrule
\end{tabular}
\vspace{-2em}
\end{table}

\section{Conclusion}
\label{sec:conclusion}
ULTRA is a unified framework for practical humanoid whole-body loco-manipulation that moves beyond reference replay toward perception- and goal-driven autonomy. It combines an RL-formulated, physics-driven retargeting policy that scales human-object MoCap into physically consistent humanoid rollouts with a distilled multimodal controller that unifies dense tracking and sparse goal specification. Experiments show improved interaction fidelity from retargeting, a student that matches tracking performance while remaining robust under distribution shift, and RL finetuning that boosts success on out-of-distribution goals. We further validate sim-to-real transfer on Unitree G1, demonstrating reliable dense tracking and sparse goal following. Overall, ULTRA points to a scalable path for versatile loco-manipulation that adapts online from realistic sensing without test-time references.

\noindent\rebuttal{\textbf{Limitations.} Future work may extend our approach to dexterous hands for interacting with non-box-shaped objects and enable more dynamic locomotion, and incorporate pretrained perception models instead of training them from scratch.}

\noindent\small{\textbf{Acknowledgments.} This work was supported in part by NSF Grants 2106825 and 2519216, the DARPA Young Faculty Award, ONR Grant N00014-26-1-2099, NIFA Award 2020-67021-32799, Toyota Research Institute, and Amazon. This work used computational resources provided through the ACCESS program and the NAIRR Pilot. We thank the Coordinated Science Laboratory and the Center for Autonomy for access to the experimental space and the G1 humanoid robot, and John M. Hart for his assistance with the robot.}

\bibliographystyle{IEEEtran}
\bibliography{references}

\clearpage
\appendix
\setcounter{topnumber}{6}
\setcounter{bottomnumber}{4}
\setcounter{totalnumber}{8}
\setcounter{dbltopnumber}{4}
\renewcommand{\topfraction}{0.95}
\renewcommand{\bottomfraction}{0.9}
\renewcommand{\dbltopfraction}{0.95}
\renewcommand{\textfraction}{0.03}
\renewcommand{\floatpagefraction}{0.85}
\renewcommand{\dblfloatpagefraction}{0.85}
\AtBeginEnvironment{tabular}{\small}
\setlength{\tabcolsep}{4pt}

\setcounter{table}{0}
\renewcommand{\thetable}{\Alph{table}}
\renewcommand*{\theHtable}{\thetable}
\setcounter{figure}{0}
\renewcommand{\thefigure}{\Alph{figure}}
\renewcommand*{\theHfigure}{\thefigure}
\setcounter{section}{0}
\renewcommand{\thesection}{\Alph{section}}
\renewcommand*{\theHsection}{\thesection}

\noindent In this appendix, we provide additional details of our ULTRA:

\begin{enumerate}
    \item Sec.~\ref{sec:demo} describes the organization of the supplementary demo video.
    \item Sec.~\ref{sec:appendix-reward} provides additional details on retargeting and the teacher policy, including the observation and reward design.
    \item Sec.~\ref{sec:appendix-student-obs} provides additional details on the student policy, covering both the distillation stage and the RL finetuning stage.
    \item Sec.~\ref{sec:appendix-sim-setup} provides additional experimental details and setups for both simulation and real-world experiments.
\end{enumerate}

\etocdepthtag.toc{mtappendix}
\etocsettagdepth{mtchapter}{none}
\etocsettagdepth{mtappendix}{subsection}
{
  \hypersetup{
    linkcolor = black
  }
  \tableofcontents
}

\subsection{Demo Videos}
\label{sec:demo}
In our \href{https://ultrahumanoid.github.io/}{webpage}, we highlight the following capabilities: (\textbf{I}) our retargeting pipeline learns a single policy from all data and performs zero-shot retargeting to unseen, rescaled trajectories and objects; (\textbf{II}) our teacher policy transfers reliably across simulators; (\textbf{III}) the all-in-one ULTRA model faithfully tracks reference motions for object interactions; (\textbf{IV}) ULTRA supports sparse object-goal commands and fine-grained, keyboard-based control, demonstrating strong generalization; and (\textbf{V}) ULTRA completes long-horizon, object-centric goals using only egocentric perception.

\subsection{Additional Details on Retargeting and Teacher Policy}
\label{sec:appendix-reward}
We provide additional details on retargeting and teacher policy training. Both follow the same general procedure. The key difference is that the retargeting policy tracks reference motions from a different embodiment, which requires a predefined key-joint mapping when constructing rewards and observations. In this appendix, we focus on the teacher policy reward and observation; This can be easily extended to the retargeting policy, which uses the same formulations, with the additional cross-embodiment link alignment applied (Table~\ref{tab:g1_smplx_correspondence}).

\noindent\textbf{Rewards.}
The teacher policy reward combines tracking terms with smoothness and regularization penalties to facilitate sim-to-real transfer for the distilled student policy. The total reward is defined as
\begin{equation}
r_{\text{teacher}} = r_{\text{track}} + \sum_i w_i \, r_i^{\text{smooth}} .
\end{equation}
Table~\ref{tab:teacher-reward-track} lists all tracking reward terms, and Table~\ref{tab:teacher-reward-smooth} summarizes all smoothness and regularization terms.

\begin{table}[t]
\centering
\small
\begin{tabular}{l c l}
\toprule
\textbf{Unitree G1 Link} & \textbf{SMPL-X Index} & \textbf{SMPL-X Body Part} \\
\midrule
left\_hip\_yaw\_link        & 1  & Left hip \\
left\_knee\_link            & 2  & Left knee \\
left\_ankle\_roll\_link     & 3  & Left ankle \\
right\_hip\_yaw\_link       & 5  & Right hip \\
right\_knee\_link           & 6  & Right knee \\
right\_ankle\_roll\_link    & 7  & Right ankle \\
torso\_link                 & 9  & Pelvis / lower torso \\
mid360\_link                & 13 & Upper torso / spine \\
left\_shoulder\_yaw\_link   & 15 & Left shoulder \\
left\_elbow\_link           & 16 & Left elbow \\
left\_wrist\_yaw\_link      & 17 & Left wrist \\
right\_shoulder\_yaw\_link  & 34 & Right shoulder \\
right\_elbow\_link          & 35 & Right elbow \\
right\_wrist\_yaw\_link     & 36 & Right wrist \\
\bottomrule
\end{tabular}
\caption{Correspondence between Unitree G1 key links and SMPL-X body indices (52-body model) used for motion retargeting.}
\label{tab:g1_smplx_correspondence}
\end{table}

\begin{table}
\centering
\caption{\textbf{Tracking reward components for the teacher policy.}
$\mathbf{p}_l,\mathbf{q},\dot{\mathbf{p}}_l,\dot{\mathbf{q}}$ denote the simulated body joint positions, joint rotation, and their velocities; $\hat{\mathbf{p}}_l,\hat{\mathbf{q}},\hat{\dot{\mathbf{p}}}_l,\hat{\dot{\mathbf{q}}}$ are the corresponding reference quantities.
$\mathbf{p}_o,\mathbf{q}_o,\dot{\mathbf{p}}_o$ denote the simulated object position, rotation (quaternion), and linear velocity; $\hat{\mathbf{p}}_o,\hat{\mathbf{q}}_o,\hat{\dot{\mathbf{p}}}_o$ are the corresponding references.
$\angle(\mathbf{q}_o,\hat{\mathbf{q}}_o)$ is the relative rotation angle and $\mathrm{huber}(\cdot)$ is the Huber loss.
$\boldsymbol{\delta}_{ij}$ denotes the palm-to-surface distance between palm point $i$ and object surface sample $j$, with weights $w_{ij}$; hats denote reference values.
$c_l\in\{0,1\}$ is a binary contact indicator for link $l$.
The scalars $k_{\cdot}$ are temperature coefficients. All reward terms are multiplied together.}
\label{tab:teacher-reward-track}
\resizebox{\columnwidth}{!}{%
\begin{tabular}{lcc}
\toprule
\textbf{Term} & \textbf{Expression} & \textbf{Weight} \\
\midrule
\rowcolor{gray!15}
\multicolumn{3}{l}{\textit{Body Tracking:}} \\
Joint position & $\exp\left(-k_p \cdot \text{mean}(\|\mathbf{p}_l - \hat{\mathbf{p}}_l\|_2^2)\right)$ & $k_p = 10.0$ \\
Joint rotation & $\exp\left(-k_r \cdot \text{mean}(\|\mathbf{q} - \hat{\mathbf{q}}\|_2^2)\right)$ & $k_r = 5.0$ \\
Body velocity & $\exp\left(-k_{pv} \cdot \text{mean}(\|\dot{\mathbf{p}}_l - \hat{\dot{\mathbf{p}}}_l\|_2^2)\right)$ & $k_{pv} = 0.1$ \\
Joint velocity & $\exp\left(-k_{rv} \cdot \text{mean}(\|\dot{\mathbf{q}} - \hat{\dot{\mathbf{q}}}\|_2^2)\right)$ & $k_{rv} = 0.001$ \\
\midrule
\rowcolor{gray!15}
\multicolumn{3}{l}{\textit{Object Tracking:}} \\
Object position & $\exp\left(-k_{op} \cdot \text{mean}(\|\mathbf{p}_o - \hat{\mathbf{p}}_o\|_2^2)\right)$ & $k_{op} = 5.0$ \\
Object rotation & $\exp\left(-k_{or} \cdot \text{huber}(\angle(\mathbf{q}_o, \hat{\mathbf{q}}_o))\right)$ & $k_{or} = 0.5$ \\
Object linear velocity & $\exp\left(-k_{opv} \cdot \text{mean}(\|\dot{\mathbf{p}}_o - \hat{\dot{\mathbf{p}}}_o\|_2^2)\right)$ & $k_{opv} = 0.1$ \\
\midrule
\rowcolor{gray!15}
\multicolumn{3}{l}{\textit{Interaction:}} \\
Palm-to-surface & $\exp\left(-k_{\text{int}} \cdot \sum_{i,j} w_{ij} \|\boldsymbol{\delta}_{ij} - \hat{\boldsymbol{\delta}}_{ij}\|_2^2\right)$ & $k_{\text{int}} = 20.0$ \\
Contact matching & $\exp\left(-k_{\text{ct}} \cdot \text{mean}(|c_l - \hat{c}_l|)\right)$ & $k_{\text{ct}} = 5.0$ \\
\bottomrule
\end{tabular}%
}
\end{table}

\begin{table}
\centering
\caption{\textbf{Smoothness and regularization rewards for the teacher policy.}
All terms are penalties with negative weights $w_i<0$.
$\mathbf{v}_{\text{base}}$ and $\boldsymbol{\omega}_{\text{base}}$ are the base linear and angular velocities.
$\mathbf{a}_t$ is the action at time $t$; $\dot{\mathbf{q}}_t$ is the joint velocity at time $t$; and $\boldsymbol{\omega}_t$ is the base angular velocity at time $t$.
$\boldsymbol{\tau}$ denotes joint torques and $\odot$ is elementwise multiplication.
$\mathbf{q}_{\text{lim}}$ and $\boldsymbol{\tau}_{\text{lim}}$ are per-joint position and torque limits.
$\mathds{1}_{\text{contact}}$ is an indicator for foot contact; $\mathbf{f}_{xy}$ and $\mathbf{f}_z$ are the horizontal and vertical components of the contact force.
$d_{\text{feet}}$ and $d_{\text{knee}}$ are the distances between the two feet and the two knees, and $\mathrm{clamp}(\cdot)$ clips the distance to the specified interval.
$\mathbf{g}_{\perp}^{\text{feet}}$ measures foot tilt relative to gravity.
$\text{ref\_stand}$ and $\text{sim\_stand}$ indicate standing phases in the reference and simulation.
$h_{\text{term}}$ is a swing-foot height term, $c$ is a clearance/contact-related term, and $g_{\text{swing}}$ gates the swing penalty, which is active only during reference indicating swing.}

\label{tab:teacher-reward-smooth}
\resizebox{\columnwidth}{!}{%
\begin{tabular}{lcc}
\toprule
\textbf{Term} & \textbf{Expression} & \textbf{Weight} \\
\midrule
\rowcolor{gray!15}
\multicolumn{3}{l}{\textit{Velocity Penalties:}} \\
Base linear velocity & $\|\mathbf{v}_{\text{base}}\|_2$ & $-0.1$ \\
Base angular velocity & $\|\boldsymbol{\omega}_{\text{base}}\|_2^2$ & $-0.01$ \\
Joint velocity & $\|\dot{\mathbf{q}}\|_2^2$ & $-0.0004$ \\
\midrule
\rowcolor{gray!15}
\multicolumn{3}{l}{\textit{Smoothness Penalties:}} \\
Action rate & $\|\mathbf{a}_t - \mathbf{a}_{t-1}\|_2$ & $-0.1$ \\
Joint velocity change & $\|\dot{\mathbf{q}}_t - \dot{\mathbf{q}}_{t-1}\|_2^2$ & $-2 \times 10^{-5}$ \\
Angular velocity change & $\|\boldsymbol{\omega}_t - \boldsymbol{\omega}_{t-1}\|_2^2$ & $-5 \times 10^{-4}$ \\
\midrule
\rowcolor{gray!15}
\multicolumn{3}{l}{\textit{Torque \& Energy Penalties:}} \\
Torque magnitude & $\|\boldsymbol{\tau}\|_2$ & $-0.001$ \\
Energy consumption & $\|\boldsymbol{\tau} \odot \dot{\mathbf{q}}\|_2$ & $-0.0001$ \\
Joint position limits & $\sum \max(0, |\mathbf{q}| - \mathbf{q}_{\text{lim}})$ & $-5.0$ \\
Joint torque limits & $\sum \max(0, |\boldsymbol{\tau}|/\boldsymbol{\tau}_{\text{lim}} - 0.95)$ & $-1.0$ \\
\midrule
\rowcolor{gray!15}
\multicolumn{3}{l}{\textit{Foot \& Stability Penalties:}} \\
Feet orientation & $\|\mathbf{g}_{\perp}^{\text{feet}}\|_2$ & $-0.35$ \\
Foot slip & $\sqrt{\|\mathbf{v}_{\text{foot}}\|_2} \cdot \mathds{1}_{\text{contact}}$ & $-0.1$ \\
Feet stumble & $\mathds{1}(\|\mathbf{f}_{xy}\| > 4|\mathbf{f}_z|)$ & $-10.0$ \\
Feet distance & $\text{clamp}(d_{\text{feet}} - [0.25, 0.65])$ & $-0.1$ \\
Knee distance & $\text{clamp}(d_{\text{knee}} - [0.25, 0.65])$ & $-0.1$ \\
Stand on feet & $\mathds{1}(\text{ref\_stand} \land \neg\text{sim\_stand})$ & $-1.0$ \\
Swing clearance & $(w_h \cdot h_{\text{term}} + w_c \cdot c) \cdot g_{\text{swing}}$ & $-0.6$ \\
Termination & $\mathds{1}_{\text{terminated}}$ & $-50.0$ \\
\bottomrule
\end{tabular}%
}
\end{table}

\noindent\textbf{Observation.}
At time step $t$, the teacher policy receives an observation vector
\begin{equation}
\boldsymbol{o}_t = \big[\boldsymbol{o}_t^{\text{sim}},\ \boldsymbol{o}_t^{\text{ref}},\ \boldsymbol{o}_t^{\Delta},\ \boldsymbol{o}_t^{\text{ig}}\big],
\end{equation}
where the four blocks correspond to simulated state, reference targets, simulation--reference residuals, and interaction-graph features, respectively. Table~\ref{tab:teacher-obs} details the observation components and variable definitions.

\begin{table}
\centering
\caption{\textbf{Teacher policy observation space.}
At time $t$, $\boldsymbol{o}_t=[\boldsymbol{o}_t^{\text{sim}},\boldsymbol{o}_t^{\text{ref}},\boldsymbol{o}_t^{\Delta},\boldsymbol{o}_t^{\text{ig}}]$.
$\boldsymbol{o}_t^{\text{sim}}$ contains simulated quantities: root height; per-body local positions $\mathbf{p}$, rotations $\mathbf{R}$ in 6D tan-norm, linear velocities $\dot{\mathbf{p}}$, angular velocities $\boldsymbol{\omega}$, contact flags $c\in\{0,1\}$; and joint states (positions $\mathbf{q}$, velocities $\dot{\mathbf{q}}$, actions $\mathbf{a}$, torques $\boldsymbol{\tau}$, and history).
$\boldsymbol{o}_t^{\text{ref}}$ contains corresponding reference targets (hats), \textit{e.g.,} body pose $(\hat{\mathbf{p}},\hat{\mathbf{R}})$ and object pose/velocity $(\hat{\mathbf{x}}_o,\hat{\mathbf{q}}_o,\hat{\mathbf{v}}_o,\hat{\boldsymbol{\omega}}_o)$.
$\boldsymbol{o}_t^{\Delta}$ stores residuals between simulation and reference (\textit{e.g.,} $\mathbf{p}-\hat{\mathbf{p}}$ and velocity errors), and $\boldsymbol{o}_t^{\text{ig}}$ stores interaction-graph features based on SDF distances between body points and the object surface and their residuals. Two observations based on next 1-frame and next 16-frame reference are concatenated, yielding total dimension 4052.}
\label{tab:teacher-obs}
\resizebox{\columnwidth}{!}{%
\begin{tabular}{llcc}
\toprule
\textbf{Category} & \textbf{Feature} & \textbf{Dimension} & \textbf{Description} \\
\midrule
\multirow{7}{*}{$\boldsymbol{o}^{\text{sim}}$}
& Root height & 1 & Root height above ground \\
& Local body positions $\mathbf{p}$ & 114 & 39 bodies $\times$ 3 (root removed) \\
& Local body rotations $\mathbf{R}$ & 234 & 39 bodies $\times$ 6 (tan-norm) \\
& Local body velocities $\dot{\mathbf{p}}$ & 117 & 39 bodies $\times$ 3 \\
& Local body angular vel. $\boldsymbol{\omega}$ & 117 & 39 bodies $\times$ 3 \\
& Contact indicators $c$ & 39 & Binary contact flags \\
& Joint states $(\mathbf{q},\dot{\mathbf{q}},\mathbf{a},\boldsymbol{\tau})$ & 145 & Proprioception + short history \\
\midrule
\multirow{2}{*}{$\boldsymbol{o}^{\text{ref}}$}
& Body reference $(\hat{\mathbf{p}},\hat{\mathbf{R}})$ & 351 & Positions (117) + rotations (234) \\
& Object reference & 13 & Pose $(\hat{\mathbf{x}}_o,\hat{\mathbf{q}}_o)$ (7) + vel. $(\hat{\mathbf{v}}_o,\hat{\boldsymbol{\omega}}_o)$ (6) \\
\midrule
\multirow{2}{*}{$\boldsymbol{o}^{\Delta}$}
& Body residuals & 585 & Pose/velocity residuals vs. reference \\
& Object residuals & 21 & Object pose/velocity residuals vs. reference \\
\midrule
\multirow{1}{*}{$\boldsymbol{o}^{\text{ig}}$}
& Interaction graph & 234 & 39 $\times$ 3 SDF distances + residuals \\
\midrule
\multicolumn{2}{l}{\textbf{Total}} & 4052 & Concatenate 2 frames (current + future) \\
\bottomrule
\end{tabular}%
}
\end{table}

\noindent\textbf{Architecture.} The teacher policy uses a three-layer MLP with hidden dimensions 1024, 1024, and 512, and ReLU activations. It follows a separate actor–critic design, producing a 29-dimensional action output. The action mean is output with no activation (\textit{i.e.,} a linear head), while the action standard deviation is fixed rather than learned and is initialized to $-2.9$. All weights use the default Xavier initialization.

\noindent\textbf{Training.} We summarize PPO hyperparameters in Table~\ref{tab:ppo-hyper}.

\begin{table}
\centering
\caption{PPO hyperparameters.}
\label{tab:ppo-hyper}
\begin{tabular}{lc}
\toprule
\textbf{Hyperparameter} & \textbf{Value} \\
\midrule
Learning rate & $2 \times 10^{-5}$ \\
Clip ratio $\epsilon$ & 0.2 \\
GAE $\lambda$ (tau) & 0.95 \\
Discount $\gamma$ & 0.99 \\
Horizon length & 32 \\
Mini-batch size & 16384 \\
Mini epochs per update & 6 \\
Entropy coefficient & 0.0 \\
Critic loss coefficient & 5.0 \\
Bounds loss coefficient & 10.0 \\
Max gradient norm & 1.0 \\
Number of parallel envs & 4096 \\
Normalize input & True \\
Normalize value & False \\
Normalize advantage & True \\
\bottomrule
\end{tabular}
\end{table}

\subsection{Additional Details on Student Policy}
\label{sec:appendix-student-obs}
\begin{table*}
\centering
\caption{\textbf{Distillation-stage loss terms.}
$\boldsymbol{\mu}$ is the student action mean from the full model forward pass, and $\boldsymbol{\mu}^{\text{prior}}$ is the action mean from a prior-only forward pass (encoder disabled).
$\boldsymbol{a}^{\text{exp}}$ denotes the expert action mean (teacher target) used for supervision.
The prior and privileged encoder output diagonal Gaussians over the latent $z$, denoted by
$p_\theta(z\mid \boldsymbol{o})=\mathcal{N}(\boldsymbol{\mu}_p,\boldsymbol{\sigma}_p^2)$ and
$q_\phi(z\mid \boldsymbol{o},\boldsymbol{o}^{\text{priv}})=\mathcal{N}(\boldsymbol{\mu}_e,\boldsymbol{\sigma}_e^2)$.
Let $\text{ep}$ be the epoch index and
$s=\mathrm{clip}\!\left(\frac{\text{ep}-500}{3000},\,0,\,1\right)$.
The cosine ramp is $g(s)=\frac{1-\cos(\pi s)}{2}$.}
\label{tab:distill-loss}
\resizebox{0.8\textwidth}{!}{%
\begin{tabular}{lcc}
\toprule
\textbf{Term} & \textbf{Definition} & \textbf{Weight / schedule} \\
\midrule
Total loss &
$\mathcal{L}=\lambda_E \mathcal{L}_E+\lambda_{\text{KL}}\mathcal{L}_{\text{KL}}+\lambda_S \mathcal{L}_S+\lambda_A \mathcal{L}_A+\lambda_G \mathcal{L}_G+\lambda_P \mathcal{L}_P$
& see below \\
\midrule
Expert imitation &
$\mathcal{L}_E=\big\|\boldsymbol{\mu}-\boldsymbol{a}^{\text{exp}}\big\|_2^2$
& $\lambda_E=1.0$ \\
\midrule
Latent KL alignment &
$\mathcal{L}_{\text{KL}}=D_{\text{KL}}\!\left(q_\phi(z)\ \|\ p_\theta(z)\right)$
& $\lambda_{\text{KL}}=0.001+(0.1-0.001)\,g(s)$ \\
\midrule
Latent smoothness &
$\mathcal{L}_S=\big\|\big(\boldsymbol{\mu}_p+\boldsymbol{\mu}_e\big)_t-\big(\boldsymbol{\mu}_p+\boldsymbol{\mu}_e\big)_{t-1}\big\|_2^2$
& $\lambda_S=0.0001+(0.001-0.0001)\,g(s)$ \\
\midrule
Auxiliary prediction &
$\mathcal{L}_A=\mathrm{MSE}\big(\hat{\boldsymbol{y}}_{\text{aux}},\boldsymbol{y}_{\text{aux}}\big)$ with mask-weighting
& $\lambda_A=1.0$ \\
\midrule
Local-goal prediction &
$\mathcal{L}_G=\mathrm{MSE}\big(\hat{\boldsymbol{g}}_{\text{local}},\boldsymbol{g}_{\text{local}}\big)$ with mask-weighting
& $\lambda_G=1.0$ \\
\bottomrule
\end{tabular}%
}
\end{table*}

\begin{table*}
\centering
\small
\setlength{\tabcolsep}{5pt}
\renewcommand{\arraystretch}{1.08}
\caption{\textbf{RL finetuning reward components}. We mask-gate goal-related terms: if a target is unobserved, we set its reward to zero and exclude it from success checks.}
\begin{tabularx}{0.8\linewidth}{l l X}
\toprule
Term & Active when & Description \\
\midrule
Goal-reaching & target visible & Reward proximity to the sampled goal (dense). \\
Progress & target visible & Reward step-to-step improvement toward the goal (clipped). \\
Terminal bonus & success on visible targets & Add a bonus when all visible thresholds are met. \\
Termination & always & Same as in Table~\ref{tab:teacher-reward-smooth}. \\
Smoothness & always & Mostly same as in Table~\ref{tab:teacher-reward-smooth} without reference related terms \\
\bottomrule
\end{tabularx}
\label{tab:reward}
\end{table*}

\noindent\textbf{Student Policy Observation.}
The student policy observation extends the student observation with multimodal inputs including object point clouds and goal phase information. The observation is structured as $\boldsymbol{o}_t^{\text{student}} = [\boldsymbol{o}^{\text{global}}, \boldsymbol{o}^{\text{cmd}}, \boldsymbol{o}^{\text{local}}, \boldsymbol{o}^{\text{proprio}}, \boldsymbol{o}^{\text{task}}, \boldsymbol{m}]$. We summarize all components in Table~\ref{tab:student-obs}.

\begin{table}
\centering
\caption{Student policy observation space.}
\label{tab:student-obs}
\resizebox{\columnwidth}{!}{%
\begin{tabular}{llcc}
\toprule
\textbf{Category} & \textbf{Feature} & \textbf{Dimension} & \textbf{Description} \\
\midrule
\multirow{2}{*}{$\boldsymbol{o}^{\text{global}}$}
& Root position residual (xy) & 2 & Horizontal position error \\
& Heading residual (yaw) & 1 & Yaw angle error \\
\midrule
\multirow{4}{*}{$\boldsymbol{o}^{\text{cmd}}$}
& End-of-episode flag & 1 & Near episode end indicator \\
& Approaching flag & 1 & Moving toward object \\
& Leaving flag & 1 & Moving away from object \\
& Time-to-go & 1 & Normalized remaining time \\
\midrule
\multirow{2}{*}{$\boldsymbol{o}^{\text{local}}$}
& IMU residual (roll, pitch) & 2 & Local orientation error \\
& Joint position residual & 29 & Joint angle error to local target \\
\midrule
\multirow{5}{*}{$\boldsymbol{o}^{\text{proprio}}$}
& Root angular velocity & 3 & Base angular velocity \\
& IMU (roll, pitch) & 2 & Current orientation \\
& Joint positions & 29 & Current joint angles \\
& Joint velocities & 29 & Scaled by 0.05 \\
& Previous action & 29 & Last action command \\
\midrule
\multirow{1}{*}{$\boldsymbol{o}^{\text{history}}$}
& Proprioceptive history & 920 & 92 dims $\times$ 10 steps \\
\midrule
\rowcolor{gray!15}
\multicolumn{2}{l}{\textit{Object Observations:}} \\
\multirow{4}{*}{$\boldsymbol{o}^{\text{task}}$}
& Object position residual & 3 & Local frame position error \\
& Object rotation residual & 6 & Tan-norm rotation error \\
& Object position & 3 & Local frame position \\
& Point cloud (PCA sampled) & 192 & 64 points $\times$ 3 in head frame \\
\midrule
\rowcolor{gray!15}
\multicolumn{2}{l}{\textit{Observation Masks:}} \\
\multirow{4}{*}{$\boldsymbol{m}$}
& Global goal mask & 3 & Keep probability for global goal \\
& Local goal mask & 31 & Keep probability for local goal \\
& Object masks & 204 & Trans(3) + rot(6) + pos(3) + pcd(192) \\
& Goal mask & 4 & Keep probability for command \\
\midrule
\multicolumn{2}{l}{\textbf{Total}} & 1496 & \\
\bottomrule
\end{tabular}%
}
\end{table}

\noindent\textbf{Student Policy Architecture.}
ULTRA enables learning a latent representation that captures task-relevant information while maintaining robustness to noise and missing modalities.

Specifically, the student is a latent-variable policy with a 64-dimensional latent $z$ and a modality-fusion transformer. Each input modality, as summarized in Table~\ref{tab:student-obs}, is first encoded into a 256-dimensional token; point-cloud perception uses a PointNet over 64 3D points that outputs a 256-dimensional feature via a point MLP and global pooling/statistics, while all other modalities are embedded with MLP encoders into the same token space. These modality tokens are fused by a lightweight transformer with token dimension 256, two layers, four attention heads, and a 1024-dimensional feed-forward network, using GELU activations, sinusoidal positional encoding, and zero dropout.

For the latent model, a prior network predicts the Gaussian parameters of $z$ from the transformer context token using two MLP heads, each mapping 256 to 128 to 64 with ReLU. During training, a privileged encoder takes the teacher observation together with additional observations and privileged signals, processes them with an MLP of widths 2048, 1024, 512, and 256, and outputs the posterior with the same 256 to 128 to 64 heads. An auxiliary latent decoder maps $z$ through a small MLP from 64 to 256 to 16 for reconstruction or regularization. Finally, $z$ conditions the policy via FiLM by predicting per-layer scale and shift parameters with a linear projection, and the FiLM modulation is scaled by 0.1.

\noindent\textbf{Distillation Training.}
ULTRA is first pretrained via a distillation loop that combines on-policy rollouts with teacher supervision. 
We apply a DAgger mixing schedule to transition from teacher-driven rollouts to student-driven rollouts. Specifically, we use full teacher rollout below 500 epoch, then linearly anneal to over epochs 1500 epoch for full student rollout. We use 4096 parallel environments with horizon length 8, mini-batch size 4096, and 2 mini-epochs per update, which updates more frequently compared to PPO setup in Table~\ref{tab:ppo-hyper}, given that supervised distillation is much more stable. The learning rate follows a warmup-and-decay schedule: it starts at $2\times 10^{-4}$, ends warmup at epoch 500, and decays to $5\times 10^{-5}$ by epoch 5500. The latent dimension is 64.

\noindent\textbf{Distillation Loss.}
In the distillation stage, the optimization uses supervised and latent-regularization objectives. The PPO actor-critic terms are disabled in the provided implementation, so the total loss is a weighted sum of (\textbf{I}) expert supervision on the action mean, (\textbf{II}) KL alignment between the privileged posterior and the student prior, (\textbf{III}) a temporal smoothness penalty on the latent mean, (\textbf{IV}) auxiliary masked-prediction losses, and (\textbf{V}) a prior-only action matching loss computed from an additional forward pass that disables the privileged encoder. We set the expert loss coefficient, auxiliary loss coefficient, and local goal prediction coefficient to 1.0. To stabilize training and encourage a useful prior, the KL coefficient increases from 0.001 to 0.1 with a cosine schedule. More details are presented in Table~\ref{tab:distill-loss}. 

\noindent\textbf{RL Finetuning Training.}
We finetune the distilled student with PPO. Importantly, PPO is applied to the \emph{deployable} student policy, \textit{i.e.,} the prior-only policy that does not use privileged encoder inputs. And to keep the prior, we preserve $1/4$ of environment still for distillation update. The PPO objective follows the standard loss. To keep finetuning stable, we use a conservative update regime. First, we scale the overall PPO contribution by a small constant relative to the distillation objectives. Second, we apply a short warm-up schedule for the critic: the policy-gradient term is gradually enabled over an initial window of $100$ training epochs, while the critic loss remains active throughout. This warm-up reduces abrupt distribution shift from the distilled policy and improves optimization stability in the early finetuning phase.

\noindent\textbf{RL Finetuning Reward.}
At each step, we define the total reward as the sum of a dense goal-reaching term, a progress term, and auxiliary regularizers: the dense term encourages matching the object and root goals via an exponentially decayed function of their current distances, while the progress term rewards step-to-step reduction in those distances, which is clipped to prevent large spikes; importantly, both goal-related terms are visibility-gated by binary masks so that if the object or root goal is hidden by the observation mask, the corresponding reward contribution is set to zero, and we add a terminal success bonus when all visible goal constraints fall below preset thresholds. More details are discussed in Table~\ref{tab:reward}.

\begin{table}
\centering
\caption{Domain randomization ranges.}
\label{tab:dr}
\begin{tabular}{llc}
\toprule
\textbf{Category} & \textbf{Parameter} & \textbf{Range} \\
\midrule
\multirow{4}{*}{Humanoid}
& Added base mass (kg) & $[-3.0, 3.0]$ \\
& CoM offset (m) & $[-0.05, 0.05]$ \\
& Motor strength scale & $[0.8, 1.2]$ \\
\midrule
\multirow{2}{*}{Environment}
& Ground friction & $[0.5, 2.0]$ \\
& Gravity perturbation (m/s$^2$) & $[-0.1, 0.1]$ \\
& Gravity randomization interval & 4 s \\
\midrule
\multirow{2}{*}{Perturbations}
& Max push velocity (m/s) & 1.0 \\
& Push interval & 4 s \\
\midrule
\multirow{1}{*}{Action}
& Action delay buffer length & 8 steps \\
\bottomrule
\end{tabular}
\end{table}

\begin{table}
\centering
\caption{Object domain randomization ranges.}
\label{tab:obj-dr}
\begin{tabular}{lc}
\toprule
\textbf{Parameter} & \textbf{Range} \\
\midrule
Object mass (kg) & $[0.15, 1.5]$ \\
Object mass rare (kg) & $[0.05, 0.15]$ (every 10 episodes) \\
Object CoM offset (m) & $[-0.05, 0.05]$ \\
Object inertia scale & $[0.5, 2.0]$ \\
Object friction & $[0.2, 1.2]$ \\
Object restitution & $[0.0, 0.3]$ \\
Object rolling friction & $[0.0, 0.05]$ \\
Object torsion friction & $[0.0, 0.05]$ \\
\bottomrule
\end{tabular}
\end{table}

\begin{table}
\centering
\caption{Point cloud domain randomization.}
\label{tab:point-dr}
\begin{tabular}{lc}
\toprule
\textbf{Parameter} & \textbf{Value} \\
\midrule
Gaussian noise std (m) & 0.02 \\
Point dropout probability & 0.15 \\
Outlier probability & 0.05 \\
Outlier max distance (m) & 0.5 \\
Depth noise scale & 0.01 \\
Density range & [0.5, 1.0] \\
Cluster noise std (m) & 0.005 \\
Scale range & [0.95, 1.05] \\
Translation noise (m) & 0.02 \\
Occlusion probability & 0.1 \\
Camera rotation noise (rad) & 0.05 \\
Camera position noise (m) & 0.02 \\
\bottomrule
\end{tabular}
\end{table}

\begin{table}
\centering
\caption{Observation noise scales.}
\label{tab:noise}
\begin{tabular}{lc}
\toprule
\textbf{Observation} & \textbf{Noise Scale} \\
\midrule
Joint positions (rad) & 0.01 \\
Joint velocities (rad/s) & 0.1 \\
Angular velocity (rad/s) & 0.05 \\
IMU & 0.05 \\
Root position (m) & 0.05 \\
\bottomrule
\end{tabular}
\end{table}

\begin{table}
\centering
\caption{Simulation configuration.}
\label{tab:sim-config}
\begin{tabular}{llc}
\toprule
\textbf{Simulation} & \textbf{Parameter} & \textbf{Value} \\
\midrule
\multirow{8}{*}{Physics}
& Simulation substeps & 1 \\
& Control frequency inverse & 17 ($\approx$59 Hz) \\
& Number of parallel envs & 4096 \\
& PhysX solver type & 1 (TGS) \\
& Position iterations & 4 \\
& Velocity iterations & 1 \\
& Contact offset (m) & 0.02 \\
& Rest offset (m) & 0.0 \\
& Bounce threshold vel. (m/s) & 0.2 \\
& Max depenetration vel. (m/s) & 1.0 \\
\midrule
\multirow{2}{*}{Ground Plane}
& Static friction & 1.0 \\
& Dynamic friction & 1.0 \\
& Restitution & 0.0 \\
\bottomrule
\end{tabular}
\end{table}

\begin{figure}
    \centering
    \includegraphics[width=\linewidth]{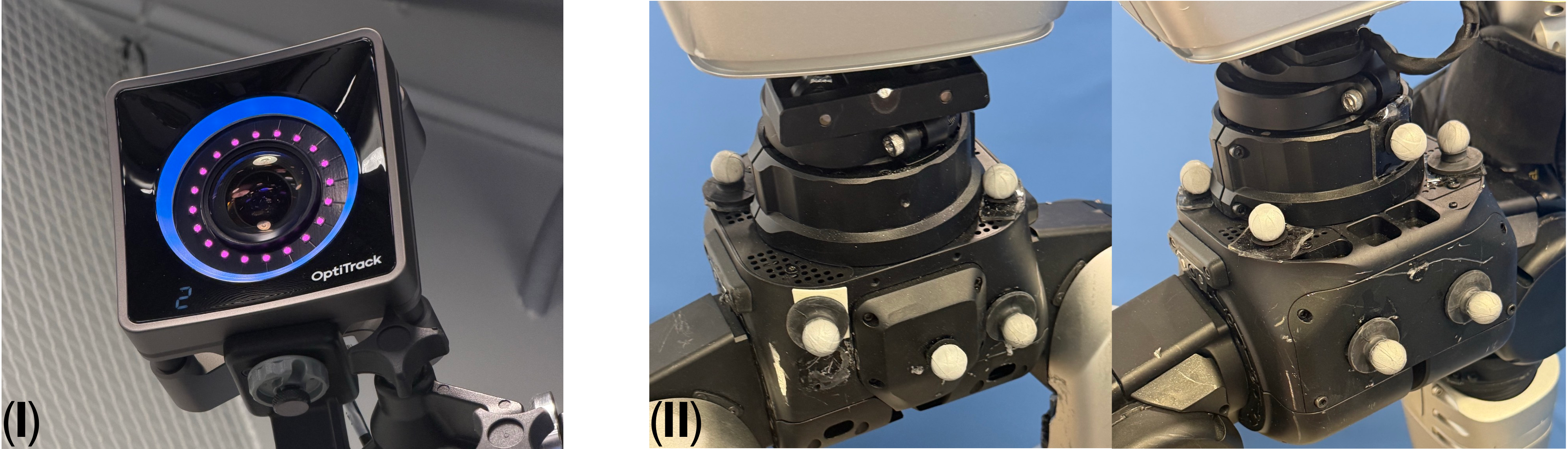}
    \caption{(\textbf{I}) The \textit{OptiTrack} camera setup used to support ULTRA control in MoCap mode. (\textbf{II}) Markers attached to the humanoid root (front and back).}
    \label{fig:mocap}
\end{figure}

\begin{figure}
    \centering
    \includegraphics[width=\linewidth]{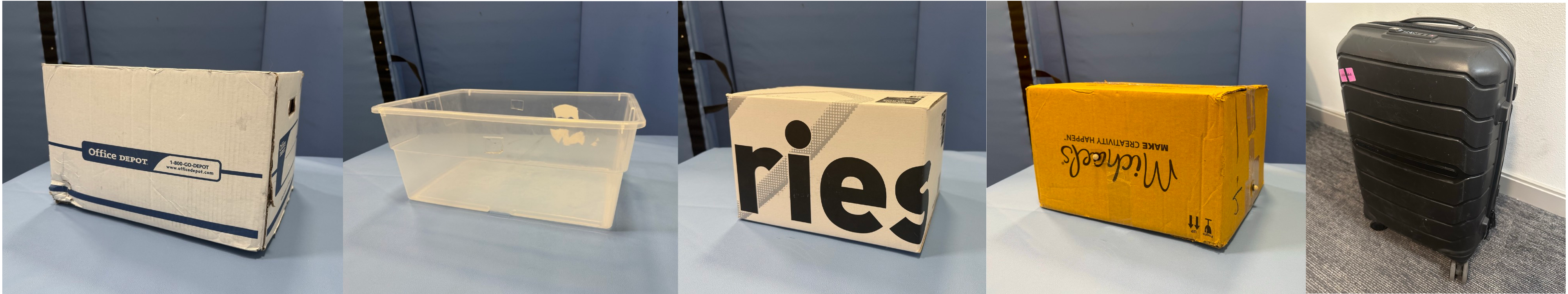}
    \caption{We visualize the objects used in our real-world deployment.}
    \label{fig:object}
\end{figure}

\subsection{Additional Experimental Details.}
\label{sec:appendix-sim-setup}
\noindent\textbf{Simulation Configuration.}
We summarize the simulation hyperparameters in Table~\ref{tab:sim-config}.

\noindent\textbf{Domain Randomization and Observation Noise.}
We summarize the domain randomization settings for the humanoid and the object in Tables~\ref{tab:dr} and~\ref{tab:obj-dr}, respectively. Observation noise is summarized in Table~\ref{tab:noise}. We additionally apply domain randomization and noise to egocentric perception, summarized in Table~\ref{tab:point-dr}.

\noindent\textbf{Real-World Deployment.}
Figure~\ref{fig:mocap} illustrates the motion-capture system used to support MoCap-driven control, and Figure~\ref{fig:object} shows the objects used in our real-world experiments.

\noindent\textbf{Limitations.}
Our method still has several limitations. It can fail under out-of-distribution conditions, such as severe point-cloud occlusion that removes critical geometric cues. In real-world experiments, MoCap-driven control is sensitive to marker occlusions, which can introduce jitter and drift in the estimated object pose and cascade into unstable tracking. Performance can also degrade when a human operator specifies overly aggressive or inconsistent goals (\textit{e.g.}, large, discontinuous target jumps or goals that are physically infeasible given the current contacts).

\end{document}